%% file: main.tex
\documentclass[10pt,twocolumn,letterpaper]{article}

\usepackage{cvpr}              % To produce the CAMERA-READY version
\usepackage{amsmath}
\usepackage{stfloats}

\input{preamble}

\definecolor{cvprblue}{rgb}{0.21,0.49,0.74}
\usepackage[pagebackref,breaklinks,colorlinks,allcolors=cvprblue]{hyperref}
\usepackage[table]{xcolor}
\usepackage{multirow}
\usepackage{tabularx}
\def\paperID{***} % *** Enter the Paper ID here
\def\confName{CVPR}
\def\confYear{2026}

\title{DUO-VSR: Dual-Stream Distillation for One-Step Video Super-Resolution}

\author{
Zhengyao Lv$^{1*}$ \hspace{0.8em}
Menghan Xia$^{2\dagger}$ \hspace{0.8em}
Xintao Wang$^{3}$ \hspace{0.8em}
Kwan-Yee K.~Wong$^{1\dagger}$\\
{\small
$^{1}$The University of Hong Kong \hspace{1.2em}
$^{2}$Huazhong University of Science and Technology \hspace{1.2em}
$^{3}$Kling Team, Kuaishou Technology 
}
}

\begin{document}

\twocolumn[{
\renewcommand\twocolumn[1][]{#1}
\maketitle

\vspace{-3.7em}
\begin{center}
{\small Project webpage: \url{https://cszy98.github.io/DUO-VSR/}}
\end{center}
\vspace{-1em}

\begin{center}
    \centering
    \vspace{-0.8em}
    \includegraphics[width=.92\linewidth]{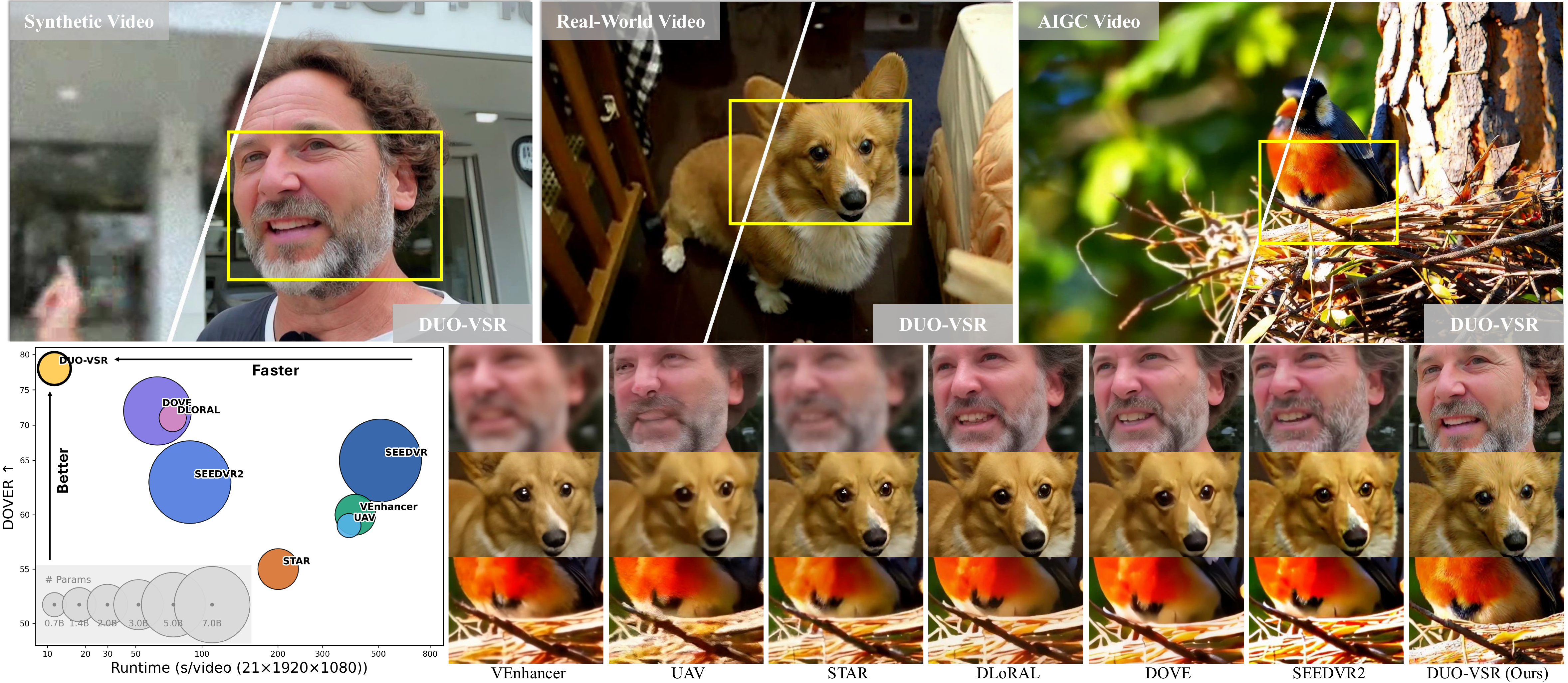}
    \vspace{-1.em}
    \captionof{figure}{Inference Speed and Performance Comparison. The bubble chart on the left compares model parameter scale, inference time, and DOVER score across methods, with inference speed measured on a single GPU using a 21-frame, $1920\times1080$ resolution video. The right-side images show super-resolution results for different videos. Our method not only demonstrates remarkable detail generation capabilities but also achieves superior inference efficiency, accelerating inference speed by approximately $50\times$ compared to SeedVR-7B.}
    \vspace{-0.3em}
    \label{fig:teaser}
\end{center}
}]

\renewcommand{\thefootnote}{}
\footnotetext{\parbox[t]{\linewidth}{\raggedright
*Work done during an internship at Kling Team, Kuaishou Tech.\\
$^{\dag}$Corresponding Author.}}

\input{sec/0_abstract}    
\input{sec/1_intro}
\input{sec/2_relatedwork}
\input{sec/3_method}
\input{sec/4_exp}

\input{sec/5_conclusion}
\input{sec/6_ack}

{
    \small
    \bibliographystyle{ieeenat_fullname}
    \bibliography{main}
}

\input{sec/X_suppl}

\end{document}

%% file: preamble.tex
\usepackage[ruled,vlined]{algorithm2e}

%% file: sec/0_abstract.tex
\begin{abstract}
Diffusion-based video super-resolution (VSR) has recently achieved remarkable fidelity but still suffers from prohibitive sampling costs. While distribution matching distillation (DMD) can accelerate diffusion models toward one-step generation, directly applying it to VSR often results in training instability alongside degraded and insufficient supervision.
To address these issues, we propose \textbf{DUO-VSR}, a three-stage framework built upon a \textbf{DU}al-Stream Distillation strategy that unifies distribution matching and adversarial supervision for \textbf{O}ne-step VSR.
Firstly, a Progressive Guided Distillation Initialization is employed to stabilize subsequent training through trajectory-preserving distillation.
Next, the Dual-Stream Distillation jointly optimizes the DMD and Real–Fake Score Feature GAN (RFS-GAN) streams, with the latter providing complementary adversarial supervision leveraging discriminative features from both real and fake score models.
Finally, a Preference-Guided Refinement stage further aligns the student with perceptual quality preferences.
Extensive experiments demonstrate that DUO-VSR achieves superior visual quality and efficiency over previous one-step VSR approaches.
\end{abstract}

%% file: sec/1_intro.tex
\vspace{-1.6em}
\section{Introduction}
\vspace{-0.3em}
Video super-resolution (VSR) aims to recover high-resolution (HR) videos from low-resolution (LR) inputs~\cite{jo2018deep,wang2019edvr}, serving as a fundamental technique for video quality enhancement.
Beyond reconstruction-based methods~\cite{chan2021basicvsr,chan2022basicvsr++,wang2019edvr}, recent studies have increasingly turned to generative paradigms~\cite{wang2021real}, particularly diffusion models~\cite{ho2020denoising,song2020score}, which offer superior visual quality and realism.
By leveraging large-scale pretrained priors~\cite{rombach2022high,yang2024cogvideox}, these models achieve remarkable detail restoration even under challenging degradations~\cite{xie2025star,wang2025turbovsr}.
Despite their impressive performance, these methods rely on dozens of iterative sampling steps~\cite{zhou2024upscale,wang2025seedvr}, which incur substantial inference computational overhead and latency, making them impractical for real-world deployment.

A common strategy to accelerate diffusion models is to reduce the number of sampling steps~\cite{song2023consistency,yin2024one}, which has been widely explored in image super-resolution~(ISR)~\cite{yue2024efficient,wu2024one,dong2025tsd}.
One line of work extends this idea to VSR by adapting one-step ISR models with temporal alignment modules~\cite{sun2025one,liu2025ultravsr}, which requires additional fine-tuning to maintain temporal consistency.
Another line of research distills pretrained multi-step text-to-video~(T2V)~\cite{yang2024cogvideox} or VSR~\cite{wang2025seedvr} models into one-step generators for VSR. DOVE~\cite{chen2025dove} stabilizes training with a regression loss, but it tends to compromise fine details. SeedVR2~\cite{wang2025seedvr2} improves perceptual fidelity via adversarial post-training~\cite{lin2025diffusion}, but it often suffers from instability due to the large discriminator which may dominate the optimization dynamics and introduce unnatural artifacts.
Despite these advances, one-step VSR methods still face trade-offs among stability, temporal consistency, and perceptual quality, thereby motivating the exploration of alternative distillation strategies.

Recently, Distribution Matching Distillation (DMD)~\cite{yin2024one,yin2025slow} has proven effective for accelerating video diffusion models, outperforming GAN-based counterparts~\cite{huang2025self}.
It trains a student model to directly match the distribution of a pretrained teacher, thereby enabling one-step generation.
However, applying DMD to VSR reveals three key limitations.
\textbf{(1) Training instability.} 
Directly initializing the student from a pretrained multi-step VSR model produces one-step outputs whose distribution deviates substantially from real HR videos, leading to instability in subsequent training.
\textbf{(2) Degraded supervision.} The frozen real score model~(i.e., the teacher model), never exposed to the noised versions of the student outputs, may produce biased or spatially shifted guidance relative to the given LR anchor, causing artifacts or temporal inconsistencies.
\textbf{(3) Insufficient supervision.} Although the real score model generates visually high-quality results, it still falls short of real HR videos, which fundamentally limits the achievable performance of the student when relying solely on DMD.

To address these issues, we introduce a three-stage distillation framework, featuring a novel \textbf{DU}al-Stream Distillation strategy that unifies distribution matching and adversarial supervision for \textbf{O}ne-step \textbf{VSR}, termed \textbf{DUO-VSR}.
We first perform Progressive Guided Distillation to obtain a one-step initialization that stabilizes subsequent training.
In the second stage, we introduce the Dual-Stream Distillation, where the distribution matching distillation stream ensures stable alignment with the teacher distribution, while the Real–Fake Score Feature GAN~(RFS-GAN) stream provides supervision from high-quality real videos.
Unlike DMD2~\cite{yin2024improved}, which applies GAN loss only during a late fine-tuning stage and computes it solely from features of the fake score model, we jointly optimize both streams and incorporate features from both real and fake score models.
The adversarial supervision from real videos serves as a regularizing signal, mitigating the adverse influence of degraded supervision from the real score model and enabling the student to achieve higher visual quality.
Finally, we apply Preference-Guided Refinement to further boost perceptual quality through preference alignment optimization.

Extensive experiments demonstrate that DUO-VSR achieves superior perceptual quality over prior one-step VSR methods.
Our main contributions are as follows:
\begin{itemize}
\item We identify the optimization challenges in applying DMD alone to one-step VSR training, namely instability and inherent degraded and insufficient supervision.
\item We propose a Dual-Stream Distillation Strategy that jointly optimizes DMD and RFS-GAN losses, alleviating the adverse effects of degraded supervision and breaking the quality bound of the teacher model.
\item We develop a three-stage pipeline with Progressive Guided Distillation, Dual-Stream Distillation, and Preference-Guided Refinement, enabling stable optimization and high-quality one-step video super-resolution.
\end{itemize}

%% file: sec/2_relatedwork.tex
\vspace{-0.6em}
\section{Related Work}
\vspace{-0.4em}
\subsection{Video Super-Resolution}
Video Super-Resolution (VSR) aims to recover high-quality videos from degraded inputs by leveraging spatial and temporal information.
Early sliding-window-based~\cite{yi2019progressive, li2020mucan}, recurrent-based~\cite{isobe2020video,chan2021basicvsr,chan2022basicvsr++, liang2022recurrent, shi2022rethinking}, as well as other VSR methods~\cite{jo2018deep,xue2019video,wang2019edvr,lucas2019generative,tian2020tdan, li2023simple,chen2024learning,liang2024vrt,youk2024fma,xu2025videogigagan}, mainly rely on synthetic degradation, which limits their applicability in real-world scenarios.
More recent efforts~\cite{yang2021real,wang2023benchmark,xie2023mitigating,zhang2024realviformer} have increasingly focused on addressing VSR in real-world scenarios. These works have explored various architectural designs~\cite{pan2021deep,wu2022animesr} and degradation pipelines~\cite{wang2021real,chan2022investigating}, yet they still struggle to synthesize realistic textures and fine details.

With the rapid advancement of diffusion models~\cite{ho2020denoising, song2020score,peebles2023scalable}, several diffusion-based VSR methods~\cite{he2024venhancer,li2025diffvsr} have demonstrated remarkable performance.
Some methods incorporate additional temporal modules into pretrained T2I models~\cite{rombach2022high,zhang2023i2vgen} to leverage their rich priors while ensuring temporal consistency.
Upscale-A-Video~\cite{zhou2024upscale} enhances a pretrained diffusion model by integrating temporal layers and a flow-guided recurrent latent propagation module.
MGLD-VSR~\cite{yang2024motion} employs a motion-guided loss to guide the diffusion process and embeds a temporal module in the decoder for temporal modeling. 
Several other works directly leverage pretrained T2V models~\cite{yang2024cogvideox,wan2025wan,bai2025vivid} for VSR. 
In STAR~\cite{xie2025star}, fine details are recovered through a local enhancement module integrated into the model. 
Besides, SeedVR~\cite{wang2025seedvr} adopts a sliding-window strategy to process long video sequences.
However, the considerable parameter scale and iterative denoising of diffusion models lead to substantial latency, hindering real-world deployment.

\vspace{-0.4em}
\subsection{Diffusion Model Acceleration}
\vspace{-0.2em}
Acceleration methods for diffusion models typically include caching-based strategies~\cite{zhao2024real,lv2024fastercache}, efficient attention~\cite{zhang2025vsa,zhang2025fast}, and distillation.
Existing distillation methods for accelerating diffusion models generally fall into two main categories, namely trajectory-preserving and distribution-matching.
Trajectory-preserving distillation exploits the ODE trajectory of diffusion models to match teacher outputs with fewer steps, as exemplified by methods such as progressive distillation~\cite{salimans2022progressive, meng2023distillation}, consistency distillation~\cite{song2023consistency,luo2023latent,kim2023consistency,ren2024hyper,wang2024phased, lu2024simplifying,lv2025dcm}, and rectified flow~\cite{liu2022flow,liu2023instaflow,yan2024perflow,ke2025flowsteer}.
Distribution-matching distillation bypasses the ODE trajectory and trains the student to align with the distribution of the teacher model. This can be achieved either through adversarial training~\cite{kang2024distilling,xu2024ufogen,luo2024you,sauer2024fast,sauer2024adversarial} or through score distillation~\cite{luo2023diff,luo2024one,zhou2024score,yin2024one,yin2024improved}.
Due to the inherent difficulty of preserving diffusion trajectories in few-step settings, trajectory-preserving methods often produce blurry results, whereas distribution-matching approaches tend to yield better video quality under few-step sampling.
Despite their effectiveness, GAN-based distribution matching~\cite{lin2025diffusion} often suffers from training instability caused by heavy discriminators, whereas DMD~\cite{yin2024one} has been widely adopted in autoregressive video generation~\cite{yin2025slow,huang2025self} for its efficiency.

\vspace{-0.4em}
\subsection{One-Step Video Super-Resolution}
\vspace{-0.4em}
Based on these acceleration methods, recent image super-resolution~(ISR) studies~\cite{yue2024efficient,wu2024one,dong2025tsd,he2024one,li2025one,noroozi2024you, sami2024hf,wang2024sinsr,xie2024addsr,zhu2024oftsr,you2025consistency} have investigated efficient few-step diffusion sampling.
In VSR, SEEDVR2~\cite{wang2025seedvr2} explores applying Adversarial Post-Training~(APT)~\cite{lin2025diffusion} to VSR, enabling one-step diffusion.
DOVE~\cite{chen2025dove} introduces a latent-pixel training strategy that employs a two-stage scheme to adapt pretrained T2V model to one-step VSR.
UltraVSR~\cite{liu2025ultravsr} introduces a degradation-aware reconstruction scheduling that reformulates multi-step denoising into a single-step process.
DLoraL~\cite{sun2025one} extends ISR-based one-step models with temporal
alignment.
Nevertheless, existing one-step methods still exhibit limited realism and temporal consistency.

%% file: sec/3_method.tex
\begin{figure*}[!t]
\centering
\includegraphics[width=0.92\linewidth]{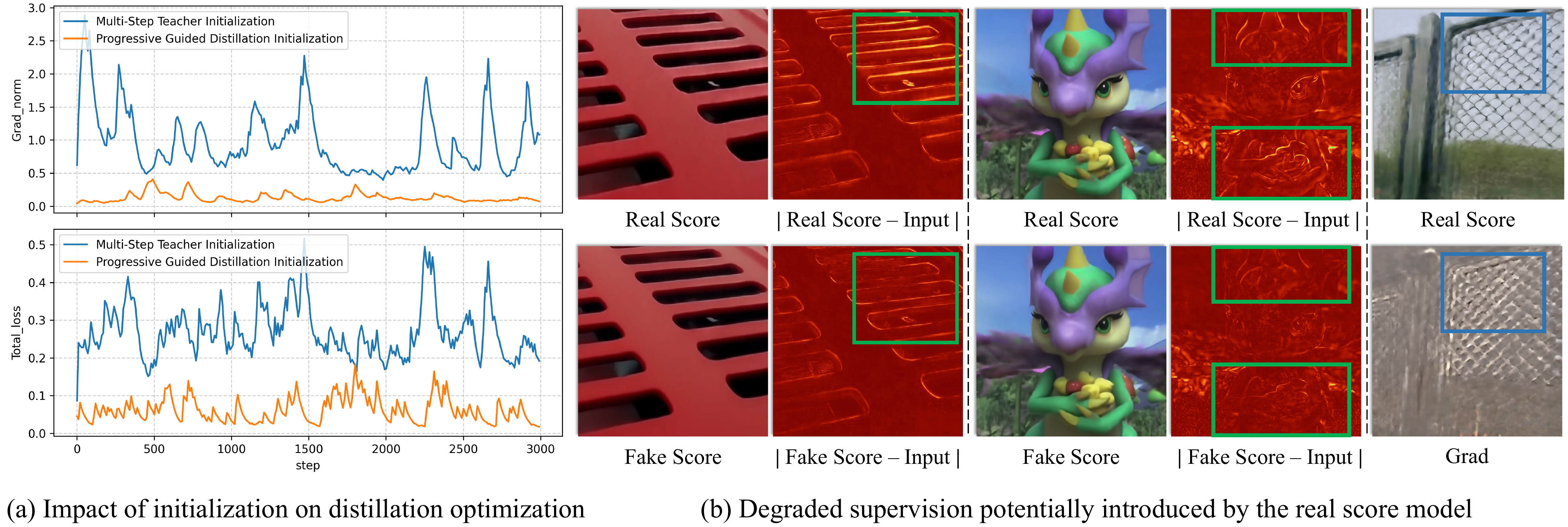}
\vspace{-0.8em}
\caption{(a) Effect of initialization on the stability of the second-stage training. The proposed progressive guided distillation initialization leads to more stable loss and gradient norm trends during the second-stage distillation. (b) Compared with the fake score model, the real score model occasionally produces outputs that are spatially shifted relative to the inputs (highlighted in \textcolor[HTML]{4FAC5B}{green} boxes in the first two cases) or contain artifacts (\textcolor[HTML]{4172B0}{blue} boxes in the third case), leading to degraded supervision propagated to the student model.}
\vspace{-1.8em}
\label{fig:motivation}
\end{figure*}
\vspace{-0.8em}
\section{Methodology}
\vspace{-0.4em}
\subsection{Preliminary}
\noindent\textbf{Base VSR Model.\ }Given a low-resolution video $x^{LR}$, we first upscale it to the target resolution and then encode it into the latent space using a VAE $\mathcal{E}$ to obtain its latent representation $\boldsymbol{z}^{LR}$. We train a video diffusion transformer~(DiT)~\cite{peebles2023scalable} conditioned on $\boldsymbol{z}^{LR}$ and text embedding $\boldsymbol{c}$, to predict clean HR latents from noisy samples obtained by perturbing HR latents $\boldsymbol{z}^{HR}$ with random noise $\epsilon$
{
\setlength{\abovedisplayskip}{3pt}
\setlength{\belowdisplayskip}{3pt}
\begin{align}
\boldsymbol{z}_t^{HR} = (1-t)\boldsymbol{z}_0^{HR} + t\epsilon, \quad
\epsilon \sim \mathcal{N}(0,I)
\end{align}
}%
\noindent where $t\in [0,1]$. The VSR denoiser $\boldsymbol{v}_{\theta}$ with parameter $\theta$ is
trained to predict the target velocity $\boldsymbol{v}=\epsilon - \boldsymbol{z}_0^{HR}$
{
\setlength{\abovedisplayskip}{3pt}
\setlength{\belowdisplayskip}{3pt}
\begin{align}
\mathcal{L}(\theta) = \mathbb{E}_{t,\boldsymbol{z}_0^{HR}}||\boldsymbol{v}_{\theta}(\boldsymbol{z}_t^{HR},t,\boldsymbol{z}^{LR},\boldsymbol{c})-\boldsymbol{v}||^2.
\label{eq:train_vsr}
\end{align}
}%

Following previous works~\cite{wang2025seedvr}, we concatenate the noisy HR latents $\boldsymbol{z}_t^{HR}$ and the LR latents $\boldsymbol{z}^{LR}$ as the input to the DiT.  
Similar to recent advanced video DiT models~\cite{yang2024cogvideox,wan2025wan}, our DiT layers incorporate a cross-attention module to integrate textual conditioning, as well as 3D full attention to capture long-range spatial and temporal dependencies. 
Our base VSR model, containing about one billion parameters, requires 50 sampling steps by default to generate clean high-resolution videos.

\vspace{-0.6em}
\noindent\textbf{Distribution Matching Distillation~(DMD).\ }DMD~\cite{yin2024one,yin2024improved} distills a multi-step diffusion model into a one-step student generator by minimizing the expected approximate Kullback-Leibler (KL) divergence $D_{KL}$ between the diffused target and student distributions over timesteps $t$.

Given a pretrained diffusion model, the distribution score can be formulated as $s = - \frac{\boldsymbol{z}_t^{HR} + (1-t)\boldsymbol{v}_{\theta}}{t}$~\cite{song2020score}, allowing the student parameters $\theta_{\text{S}}$ to be optimized by directly computing the gradient of the KL divergence
{
\setlength{\abovedisplayskip}{3pt}
\setlength{\belowdisplayskip}{3pt}
\begin{align}
\nabla_{\theta}D_{KL} = \mathbb{E}_{\epsilon}[-(s_{\text{real}}(\boldsymbol{z}_t^{HR})-s_{\text{fake}}(\boldsymbol{z}_t^{HR}))\frac{d\boldsymbol{v}}{d\theta_{\text{S}}}],
\label{eq:dmd}
\end{align}
}%
where $s_{\text{real}}$ and $s_{\text{fake}}$ are computed by the real and fake score models, respectively. Both models are initialized with the same architecture and weights as the teacher model. The real score model is frozen during training to capture the teacher distribution, while the fake score model is continuously updated to track the student distribution.

\begin{figure*}[t]
\centering
\includegraphics[width=0.92\linewidth]{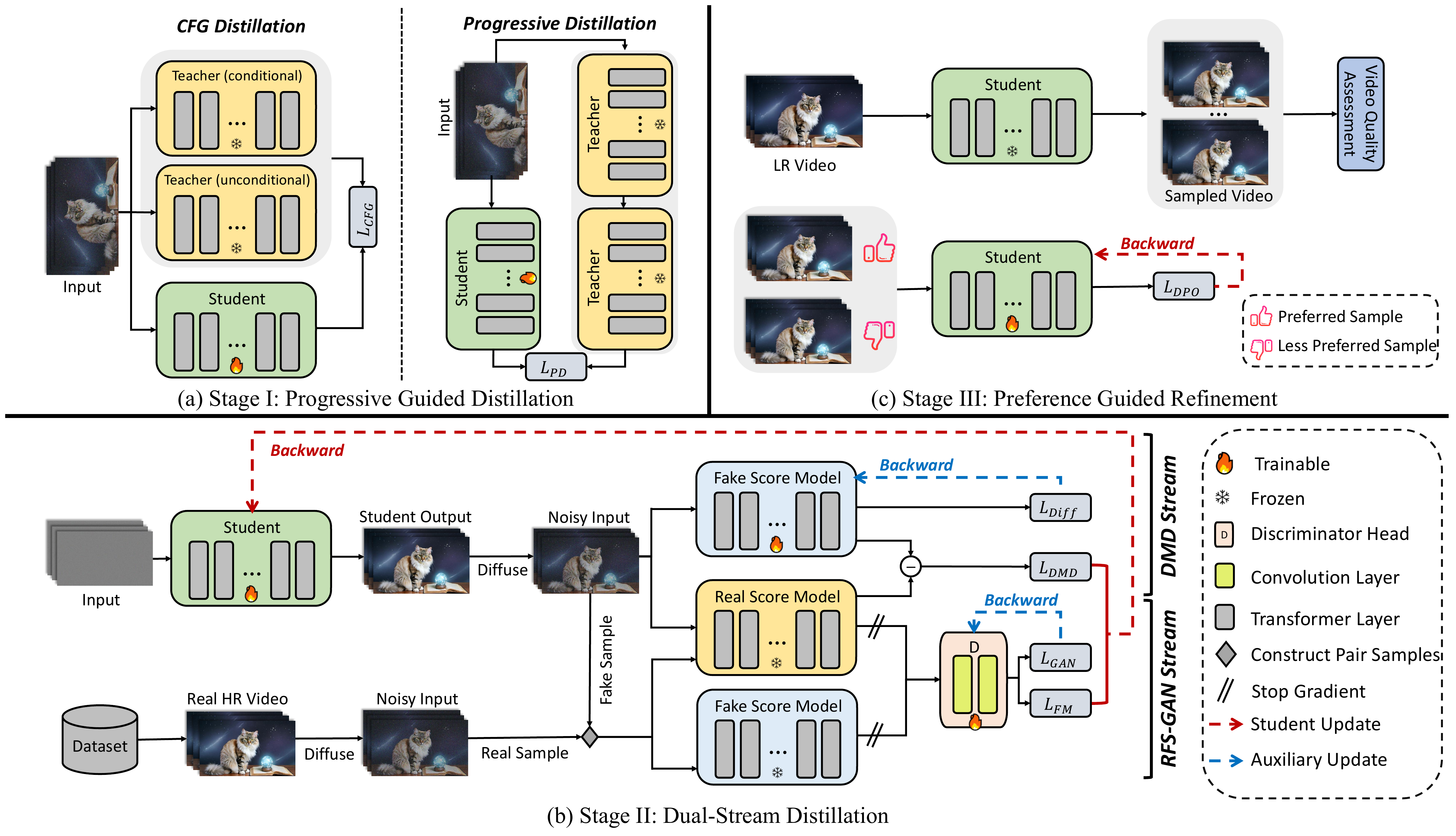}
\vspace{-1em}
\caption{Overview of our three-stage distillation framework. (a) We initialize the student model with trajectory-preserving Progressive Guided Distillation, which consists of CFG Distillation and Progressive Distillation steps. (b) The core of our method, Dual-Stream Distillation, jointly optimizes the DMD and RFS-GAN streams through alternating Student Update and Auxiliary Update, providing reliable and sufficient supervision. (c) In the final stage, we construct a generated preference dataset and apply DPO-based Preference-Guided Refinement to enhance perceptual quality.}
\vspace{-1.6em}
\label{fig:method}
\end{figure*}

\vspace{-0.8em}
\subsection{DMD in VSR: On Stability and Supervision}
\vspace{-0.8em}
Despite the impressive performance of DMD in image and video generation, we observe that directly applying it to one-step VSR training faces several challenges.
First, DMD initializes the student, real score, and fake score models from the pretrained multi-step VSR model. Since the pretrained model yields low-quality results under the one-step setting, its distribution differs notably from that of the real score model, causing unstable optimization and degraded results. As shown in Fig.~\ref{fig:motivation}~(a), directly initializing from the teacher model results in unstable gradients and training dynamics.
Second, the real score model has never been exposed to the noisy outputs of the student model.
Compared with the fake score model that continuously tracks the outputs of the student model, the real score model generates results with richer high-frequency details and textures, but often exhibit undesired spatial shifts relative to the inputs. 
Moreover, it occasionally produces artifact-contaminated outputs, which can be further propagated to the student through gradient updates, as illustrated in Fig.~\ref{fig:motivation}~(b). 
These issues are particularly evident in VSR, where the LR video serves as a strong spatial–temporal anchor, making the system more sensitive to degraded supervision than in text-conditioned image or video generation.
Finally, while the real score model represents a high-quality distribution, it remains inferior to real HR videos.
Consequently, relying solely on the DMD loss restricts the student to limited representational capacity of the teacher model.

To mitigate instability during training, we propose the Progressive Guided Distillation Initialization in Sec.~\ref{PGDI}.
To alleviate the adverse effects of degraded and insufficient supervision, we introduce the Dual-Stream Distillation Strategy in Sec.~\ref{TSDS}.
Finally, to further enhance the perceptual quality of the generated videos, we incorporate a Preference-Guided Refinement stage in Sec.~\ref{PGR}.

\vspace{-0.8em}
\subsection{Progressive Guided Distillation Initialization}
\vspace{-0.4em}
\label{PGDI}
Having identified the instability caused by direct one-step distillation, we adopt a trajectory-preserving Progressive Guided Distillation Initialization to provide a stable foundation for subsequent dual-stream optimization.

Specifically, following \citep{meng2023distillation}, we first train a single model $\theta_{\text{S}}$ to match the combined output of the conditional and unconditional diffusion branches~(CFG-Distillation in Fig.~\ref{fig:method}~(a)). This can be formulated as
{
\setlength{\abovedisplayskip}{3pt}
\setlength{\belowdisplayskip}{3pt}
\begin{align}
&\boldsymbol{v}_{\text{cfg}} = (1+w)\boldsymbol{v}_{\theta}(\boldsymbol{z}_t^{HR},t,\boldsymbol{z}^{LR},\boldsymbol{c}) - \boldsymbol{v}_{\theta}(\boldsymbol{z}_t^{HR},t,\boldsymbol{z}^{LR},\emptyset) \notag \\
&\mathcal{L}_{CFG}(\theta_{\text{S}}) = \mathbb{E}_{t,\boldsymbol{z}_0^{HR}}||\boldsymbol{v}_{\theta_{\text{S}}}(\boldsymbol{z}_t^{HR},t,\boldsymbol{z}^{LR},\boldsymbol{c})-\boldsymbol{v}_{\text{cfg}}||^2.
\label{eq:cfg}
\end{align}
}%

We then treat the CFG-Distilled $\boldsymbol{v}_{\theta_{\text{S}}}$ as the teacher model and progressively distill it into a one-step student~(Progressive Distillation in Fig.~\ref{fig:method}~(a)):
{
\setlength{\abovedisplayskip}{3pt}
\setlength{\belowdisplayskip}{3pt}
\begin{equation}
\mathcal{L}_{PD}(\theta_{\text{S}})
= \mathbb{E}_{t,\boldsymbol{z}_0^{HR}}
\big\|
\underbrace{\boldsymbol{z}_t^{HR} - (t-t'')\boldsymbol{v}_{\theta_{\text{S}}}(\boldsymbol{z}_t^{HR})}_{\text{student}}
-
\underbrace{\tilde{\boldsymbol{z}}_{t''}^{HR}(\theta)}_{\text{teacher}}
\big\|^2,
\label{eq:pd}
\end{equation}
}%
where $\boldsymbol{v}_{\theta_{\text{S}}}(\boldsymbol{z}_t^{HR})$ is the predicted velocity of student model at timestep $t$, and $\tilde{\boldsymbol{z}}_{t''}^{HR}(\theta)$ denotes the two-step prediction at timestep $t''$ obtained by integrating the teacher model over timesteps $(t, t', t'')$. For simplicity, the conditions like text embeddings and timesteps are omitted from the notations.

% \vspace{-0.6em}
\subsection{Dual-Stream Distillation Strategy}
% \vspace{-0.4em}
\label{TSDS}
Building upon a stable initialization, we further address the degraded and insufficient supervision in DMD by introducing a Dual-Stream Distillation Strategy that unifies distribution matching (DMD Stream) and adversarial supervision (RFS-GAN Stream), as shown in Fig.~\ref{fig:method}~(b).

\noindent\textbf{DMD Stream.\ }For the distribution matching distillation stream, we follow the setting in~\citep{yin2024one} and initialize both the real score model $\theta_{\text{R}}$ and fake score model $\theta_{\text{F}}$ from the pretrained teacher model.
The real score model remains frozen to capture the distribution of high-quality videos, while the fake score model is updated to track the evolving distribution of the one-step student.
During training, we optimize the fake score model $\theta_{\text{F}}$ with diffusion loss $\mathcal{L}_{Diff}$: 
{
\setlength{\abovedisplayskip}{3pt}
\setlength{\belowdisplayskip}{3pt}
\begin{align}
& \hat{\boldsymbol{z}}_0^{S} = \epsilon - \boldsymbol{v}_{\theta_{\text{S}}}(\epsilon,t,\boldsymbol{z}^{LR},\boldsymbol{c}) \notag \\
& \mathcal{L}_{Diff}(\theta_{\text{F}}) = \mathbb{E}_{t,\hat{\boldsymbol{z}}_0^{S}}||\boldsymbol{v}_{\theta_{\text{F}}}(\hat{\boldsymbol{z}}_t^{S},t,\boldsymbol{z}^{LR},\boldsymbol{c})-\boldsymbol{v}||^2,
\label{eq:streamdmd1}
\end{align}
}%
where $\hat{\boldsymbol{z}}_0^{S}$ represents the latent of the HR video predicted by one-step student model, and $\hat{\boldsymbol{z}}_t^{S}$ is obtained by diffusing it. 
Meanwhile, we alternately optimize the student model $\theta_{\text{S}}$ using the DMD loss $\mathcal{L}_{DMD}$:
{
\setlength{\abovedisplayskip}{3pt}
\setlength{\belowdisplayskip}{3pt}
\begin{align}
&\text{Grad} = \frac{\hat{\boldsymbol{z}}_0^{F}(\hat{\boldsymbol{z}}_t^{S};\theta_{\text{F}}) - \hat{\boldsymbol{z}}_0^{R}(\hat{\boldsymbol{z}}_t^{S};\theta_{\text{R}})}{\operatorname{mean}(\operatorname{abs}(\hat{\boldsymbol{z}}_0^{S}-\hat{\boldsymbol{z}}_0^{R}(\hat{\boldsymbol{z}}_t^{S};\theta_{\text{R}})))} \notag \\
&\mathcal{L}_{DMD}(\theta_{\text{S}}) = \mathbb{E}_{t,\hat{\boldsymbol{z}}_0^{S}}||\hat{\boldsymbol{z}}_0^{S}-\operatorname{sg}[\hat{\boldsymbol{z}}_0^{S}-\text{Grad}]||^2,
\label{eq:streamdmd2}
\end{align}
}%
where $\hat{\boldsymbol{z}}_0^{R}$ and $\hat{\boldsymbol{z}}_0^{F}$ are the outputs of the real and fake score models, corresponding to the real and fake scores respectively, and $\operatorname{sg}(.)$ is the stop-gradient operator.

\noindent\textbf{RFS-GAN Stream.\ }In the Real–Fake Score Feature~(RFS-GAN) stream, we employ both the frozen Real Score and the Fake Score models as discriminator backbone to extract features. 
The backbone takes the diffused output of the one-step student $\hat{\boldsymbol{z}}_t^{S}$ as fake samples and the diffused HR video $\boldsymbol{z}^{HR}_t$ as real samples, sharing the same conditioning inputs as the DMD stream, including LR video and corresponding timestep.
The intermediate features from transformer layers are concatenated and fed into additional convolutional discriminator heads to compute the RFS-GAN loss, adopting a hinge GAN objective for stable training:
{
\setlength{\abovedisplayskip}{3pt}
\setlength{\belowdisplayskip}{3pt}
\begin{align}
\mathcal{L}_{D}&=\mathbb{E}[\operatorname{max}(0,1-D(\boldsymbol{z}^{HR}_t))] + \mathbb{E}[\operatorname{max}(0,1+D(\hat{\boldsymbol{z}}_t^{S}))] \notag \\
\mathcal{L}_{G}&=-\mathbb{E}[D(\hat{\boldsymbol{z}}_t^{S})] .
\label{eq:gan}
\end{align}
}
To further stabilize training, we introduce a feature matching loss $\mathcal{L}_{FM}$ computed as the mean squared error between intermediate features extracted from the score models.

\noindent\textbf{Dual-Stream Joint Optimization.\ }To exploit the complementary strengths of DMD and adversarial supervision, we perform dual-stream joint optimization over the student, fake score model, and convolutional discriminator heads, as illustrated in Fig.~\ref{fig:method}~(b).
We alternate between two interleaved optimization phases: \textbf{(a) Student update}, where the one-step student is updated jointly by the $\mathcal{L}_{DMD}$, $\mathcal{L}_{G}$, and $\mathcal{L}_{FM}$ losses; and \textbf{(b) Auxiliary update}, where the fake score model and discriminator heads are separately updated with diffusion loss $\mathcal{L}_{Diff}$ and GAN objective $\mathcal{L}_{D}$.
We apply a stop-gradient between backbone features and discriminator heads to prevent GAN gradients from affecting the score models during discriminator head updates.
The detailed algorithm is provided in the supplementary material.

This joint formulation constitutes the core of our framework 
and delivers two interrelated benefits.
\textbf{(1) Reliable and comprehensive supervision.}
The RFS-GAN stream regularizes and complements the degraded and insufficient DMD supervision.
It suppresses the biased gradients induced when the frozen real score model encounters unseen noisy student outputs, and introduces real-video adversarial signals that enrich and extend the guidance beyond the teacher distribution.
By leveraging features from both real and fake score models, the adversarial supervision becomes more complete and balanced.
\textbf{(2) Stability and efficiency.} 
Operating on diffused samples, RFS-GAN naturally benefits from shared partial forward passes with the DMD stream, improving computational efficiency while the injected noise stabilizes adversarial dynamics.
In addition, the stop-gradient between the score-model backbones and discriminator heads decouples their optimization, ensuring that adversarial gradients do not interfere with the distribution tracking of score models.
Together, it enables a steady, efficient, and well-regularized joint training process that integrates the strengths of both streams.

\vspace{-0.6em}
\subsection{Preference-Guided Refinement}
\vspace{-0.4em}
\label{PGR}
To further enhance the perceptual quality of the one-step VSR student, we introduce a Preference-Guided Refinement, as illustrated in Fig.~\ref{fig:method}~(c).
The second-stage student model generates multiple HR candidates for each LR video, which are ranked by video quality assessment models to form a synthetic preference dataset $\mathcal{D}=\{\boldsymbol{z}^{LR},\hat{\boldsymbol{z}}^{Sw}_0,\hat{\boldsymbol{z}}^{Sl}_0\}$, with $\hat{\boldsymbol{z}}^{S_w}_0$ preferred over $\hat{\boldsymbol{z}}^{S_l}_0$.
The student model is then fine-tuned with Direct Preference Optimization (DPO)~\cite{liu2025improving} loss $\mathcal{L}_{DPO}$ to better align with perceptual preferences:
{
\setlength{\abovedisplayskip}{3pt}
\setlength{\belowdisplayskip}{3pt}
\begin{align}
&-\mathbb{E}[\operatorname{log\sigma}(-\frac{\beta_t}{2}(||\boldsymbol{v}^{w}-\boldsymbol{v}_{\theta_{\text{S}}}(\hat{\boldsymbol{z}}^{S_w}_t)||^2-||\boldsymbol{v}^{w}-\boldsymbol{v}_{\theta_{\text{ref}}}(\hat{\boldsymbol{z}}^{S_w}_t)||^2 \notag \\ &- (||\boldsymbol{v}^{l}-\boldsymbol{v}_{\theta_{\text{S}}}(\hat{\boldsymbol{z}}^{S_l}_t)||^2-||\boldsymbol{v}^{l}-\boldsymbol{v}_{\theta_{\text{ref}}}(\hat{\boldsymbol{z}}^{S_l}_t)||^2) ))],
\label{eq:dpo}
\end{align}
}%
where $\beta_t$ is a hyperparameter and $\boldsymbol{v}_{\theta_{\text{ref}}}$ is the reference model. This loss encourages $\boldsymbol{v}_{\theta_{\text{S}}}$ to approach the target velocity $\boldsymbol{v}^{w}$ of the preferred data, while repelling it from $\boldsymbol{v}^{l}$ associated with the less preferred data.
This refinement further aligns the one-step generator with perceptual preferences, yielding high-fidelity video results.

%% file: sec/4_exp.tex
% \vspace{-4em}
\section{Experiments}
\vspace{-0.4em}
\begin{figure*}[!t]
\centering
\includegraphics[width=0.85\linewidth]{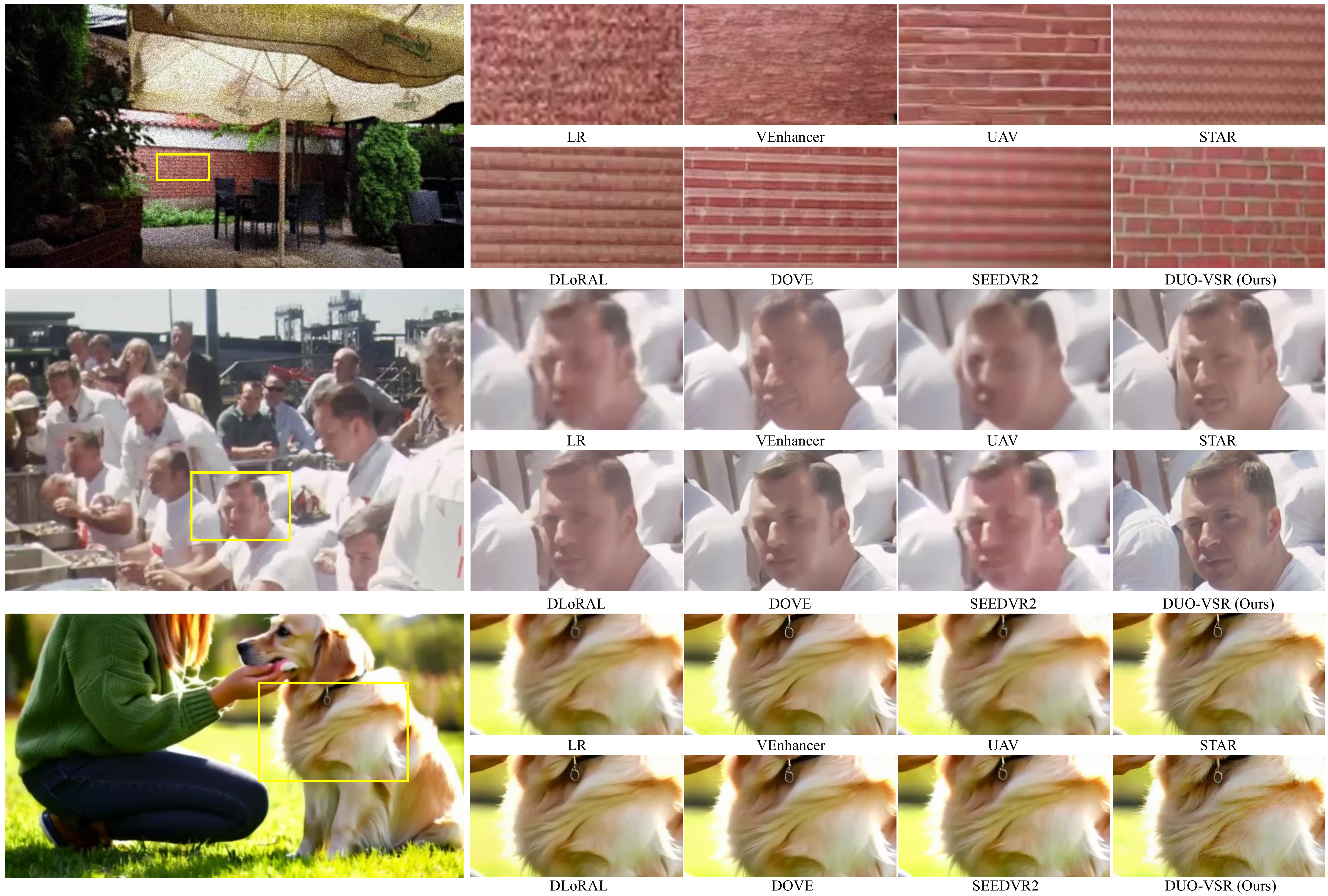}
\vspace{-1.0em}
\caption{Visual comparison on synthetic~(YouHQ40), real-world~(VideoLQ) and AIGC~(AIGC60) datasets. Zoom in for details.}
\vspace{-1.4em}
\label{fig:visualres}
\end{figure*}

\begin{table*}[!t]
\caption{Quantitative comparisons on benchmarks, including synthetic (SPMCS~\cite{tao2017detail}, UDM10~\cite{yi2019progressive}, YouHQ40~\cite{zhou2024upscale}), real-world (VideoLQ~\cite{chan2022investigating}), and AIGC (AIGC60) videos. The best and second performances are marked in {\color[HTML]{F94848} red} and {\color[HTML]{3166FF} blue} respectively.}
\vspace{-10pt}
\renewcommand\arraystretch{1.1}
\footnotesize
\begin{tabular}{c|c|ccccccccc}
\toprule
Datasets & Metrics & RealViformer & VEnhancer & ~~MGLD~~ & ~~UAV~~ & ~STAR~ & DLoRAL & ~DOVE~ & SeedVR2-7B & DUO-VSR \\ \hline \hline
           & PSNR $\uparrow$     & 21.34 & 19.92 & 21.89 & 19.67 & 21.56 & 22.87 & 22.69 & \color[HTML]{F94848}{23.08} & \color[HTML]{3166FF}{22.90} \\
           & SSIM $\uparrow$     & 0.601 & 0.523 & 0.642 & 0.538 & 0.613 & 0.674 &  \color[HTML]{F94848}{0.694} & 0.685 &  \color[HTML]{3166FF}{0.691} \\
           & LPIPS $\downarrow$  & 0.394 & 0.417 & 0.348 & 0.424 & 0.386 & 0.331 & 0.318 & \color[HTML]{F94848}{0.302} & \color[HTML]{3166FF}{0.315} \\ \cline{2-11} 
           
           & NIQE $\downarrow$   & 4.67 & 4.36 & \color[HTML]{3166FF}{3.79} & 3.87 & 6.17 & 3.96 & 4.79 & 4.68 &  \color[HTML]{F94848}{3.59} \\
           & MUSIQ $\uparrow$    & 62.13 & 61.10 & 63.43 & 58.52 &29.08  & 61.42 & \color[HTML]{3166FF}{65.27} & 64.74 & \color[HTML]{F94848}{66.91} \\
           & CLIP-IQA $\uparrow$ & 0.3443 & 0.3398 & 0.4333 & 0.4582 & 0.3827 & \color[HTML]{3166FF}{0.5218} & 0.4922 & 0.5073 & \color[HTML]{F94848}{0.5459} \\
           & DOVER $\uparrow$    & 61.32 & 58.25 & 75.54 & 69.84 & 35.17 & 72.30 & \color[HTML]{3166FF}{79.94} & 75.40 & \color[HTML]{F94848}{81.47} \\
\multirow{-9}{*}{SPMCS} & $E_{warp}^{*}\downarrow$ & 4.45 & 3.92 & 4.04 & 4.99 & 6.53 & 3.45 & \color[HTML]{3166FF}{2.10} & 2.96 & \color[HTML]{F94848}{1.67} \\ \hline
           & PSNR $\uparrow$     & 23.75 & 23.38 & 23.89 & 23.16 & 23.97 & \color[HTML]{3166FF}{24.83} & 24.32 & 24.56 & \color[HTML]{F94848}{24.94} \\
           & SSIM $\uparrow$     & 0.638 & 0.612  &  0.667 & 0.607 & 0.659  & \color[HTML]{3166FF}{0.739} & 0.723 & \color[HTML]{F94848}{0.745} & 0.726 \\
           & LPIPS $\downarrow$  & 0.364 & 0.398 & 0.385 & 0.401 & 0.324 & 0.272 & 0.284 & \color[HTML]{3166FF}{0.267} & \color[HTML]{F94848}{0.259} \\ \cline{2-11} 
           & NIQE $\downarrow$   & 5.17 & 4.99 & 4.93 & 4.64 & 5.79 & 4.65 & 4.51 & \color[HTML]{3166FF}{4.46} & \color[HTML]{F94848}{4.07} \\
           & MUSIQ $\uparrow$    & 59.11 & 54.51 & \color[HTML]{3166FF}{58.71} & 59.53 & 48.21 & 56.96 & 54.51 & 51.89 & \color[HTML]{F94848}{62.25} \\
           & CLIP-IQA $\uparrow$ & 0.4134 & 0.3859 & 0.4047 & 0.4002 & 0.2636 & 0.4163 & \color[HTML]{3166FF}{0.4412} & 0.4346 & \color[HTML]{F94848}{0.4898} \\
           & DOVER $\uparrow$    & 70.51 & 73.48 & 71.35 & 65.72 & 56.72 & 62.03 & \color[HTML]{3166FF}{74.87} & 69.09 & \color[HTML]{F94848}{75.32} \\
\multirow{-9}{*}{UDM10} & $E_{warp}^{*} \downarrow$ &  3.43 & 3.65 & 3.81 & 3.76 & 2.96 & 3.89 & \color[HTML]{3166FF}{2.89} & 3.12 & \color[HTML]{F94848}{2.44}  \\ \hline
           & PSNR $\uparrow$     & 20.98 & 19.23 & 21.35 & 18.97 & 21.76 &  22.57 & \color[HTML]{F94848}{23.12} & 22.87 & \color[HTML]{3166FF}{22.96} \\
           & SSIM $\uparrow$     & 0.621 & 0.543 & 0.661 & 0.564 & 0.642 & 0.658 & \color[HTML]{3166FF}{0.682} & \color[HTML]{F94848}{0.691} & 0.674 \\
           & LPIPS $\downarrow$  & 0.379 & 0.426 & 0.355 & 0.433 & 0.364 & 0.311 & \color[HTML]{3166FF}{0.299} & 0.291 & \color[HTML]{F94848}{0.289} \\ \cline{2-11} 
           & NIQE $\downarrow$   & 5.30 & 4.97 & 4.52 &\color[HTML]{3166FF}{4.11}  & 5.69 &4.18  & 4.91 & 4.86 & \color[HTML]{F94848}{3.92}  \\
           & MUSIQ $\uparrow$    & 54.86 & 56.83 & 57.30 & 57.52 & 49.19 & 59.64 & \color[HTML]{3166FF}{61.36} & 58.44 & \color[HTML]{F94848}{65.24} \\
           & CLIP-IQA $\uparrow$ & 0.3185 & 0.3203 & 0.4106 & 0.4028 & 0.2871 &\color[HTML]{3166FF}{0.4188}  & 0.4167 & 0.3736 & \color[HTML]{F94848}{0.4222} \\
           & DOVER $\uparrow$    & 71.13 & 75.36 & 81.23 & 83.68 & 50.96 & 68.32 & \color[HTML]{3166FF}{84.43} & 73.60 & \color[HTML]{F94848}{87.28} \\
\multirow{-9}{*}{YouHQ40} & $E_{warp}^{*}\downarrow$ & 3.63 & 3.54 & 3.32 & 2.68 & \color[HTML]{3166FF}{2.09} & 2.71 & 2.37 & 2.54 & \color[HTML]{F94848}{1.98} \\ \hline \hline
           & NIQE $\downarrow$& 5.52 & 5.38 & 4.78 & 4.77 & 5.16 & 5.17 & \color[HTML]{3166FF}{4.43} & 4.63 & \color[HTML]{F94848}{4.08} \\
           & MUSIQ $\uparrow$    & 49.20 & 46.21 & 50.87 & 51.47 & 45.90 & \color[HTML]{3166FF}{59.08} & 51.25 & 55.45 & \color[HTML]{F94848}{59.24} \\
           & CLIP-IQA $\uparrow$ & 0.3221 & 0.3106 &0.3633  & 0.3460 & 0.2753 & \color[HTML]{F94848}{0.4068} & 0.3209 & 0.3387 & \color[HTML]{3166FF}{0.3925} \\
           & DOVER $\uparrow$    & 61.09 & 58.87 & 65.36 & 62.57 & 63.43 & 69.29 & \color[HTML]{3166FF}{69.36} & 59.56 & \color[HTML]{F94848}{69.71} \\
\multirow{-5}{*}{VideoLQ} & $E_{warp}^{*}\downarrow$ & 5.03 & 4.93 & 4.10 &  4.82& 4.29 & 4.46 & \color[HTML]{3166FF}{3.91} & 4.08 & \color[HTML]{F94848}{3.67} \\ \hline
           & NIQE $\downarrow$   & 6.61 & 5.86 & 5.56 & 5.73 & 5.28 &5.44& 5.47 & \color[HTML]{3166FF}{4.99} & \color[HTML]{F94848}{4.42} \\
           & MUSIQ $\uparrow$    & 50.99 & 50.85 & 53.82 & 52.34 & 57.52 & 58.65 &57.89  & \color[HTML]{3166FF}{62.30} & \color[HTML]{F94848}{63.68} \\
           & CLIP-IQA $\uparrow$ & 0.3464 & 0.3933 & 0.4203 &  0.4209 & 0.3509 & \color[HTML]{3166FF}{0.4668} & 0.4061 & 0.4376 & \color[HTML]{F94848}{0.4886} \\
           & DOVER $\uparrow$ & 84.55  & 83.65 & 83.52 & 83.50 & 87.32 & \color[HTML]{3166FF}{87.89} & 87.61  & 86.79 & \color[HTML]{F94848}{88.15} \\
\multirow{-5}{*}{AIGC60} & $E_{warp}^{*}\downarrow$ & 3.89 & 3.76 & 3.40 & 4.17 & 1.22 & 1.67 & \color[HTML]{F94848}{0.95} & 1.48 & \color[HTML]{3166FF}{1.08} \\ 
\bottomrule
\end{tabular}
\vspace{-2.4em}
\label{tab:quantitative-results}
\end{table*}

\subsection{Experimental Settings}
\vspace{-0.4em}
\label{expsetting}
\noindent\textbf{Implementation Details.\ }Our base VSR model is built upon an internal 1.3B-parameter text-to-video model, which is adapted through 10k iterations of training on 830k paired samples synthesized by RealBasicVSR~\cite{chan2022investigating} degradation pipeline, with a batch size of 64.
In the Progressive Guided Distillation stage, we first perform CFG Distillation for 500 iterations. 
Next, starting from a 64-step teacher, we progressively halve number of denoising steps of student, using a learning rate of $5\times10^{-5}$ and a batch size of 32. Meanwhile, teacher is updated with the latest student every 500 iterations, until obtaining a single-step model.
In the Dual-Stream Distillation stage, we perform one student update after every three auxiliary updates, iterating for 2,000 steps in total. The DMD loss, RFS-GAN loss, and feature matching loss are weighted by 1.0, 0.1, and 0.05 respectively. The learning rate and batch size are set to $5\times10^{-6}$ and 32 respectively.
In the Preference-Guided Refinement stage, we construct 2,000 preference pairs and fine-tune the model for 1,000 iterations with a learning rate of $1\times10^{-6}$.

\noindent\textbf{Evaluation Settings.\ }Following previous work~\cite{wang2025seedvr}, we conduct evaluations on synthetic benchmarks including SPMCS~\cite{yi2019progressive}, UDM10~\cite{tao2017detail}, and YouHQ40~\cite{zhou2024upscale} under the same degradation settings as in training.
Furthermore, we evaluate on a real-world dataset VideoLQ~\cite{chan2022investigating} and a self-constructed AIGC60 dataset comprising 60 AI-generated videos covering a wide range of visual scenes.

For synthetic datasets, we evaluate the fidelity using full-reference metrics including PSNR, SSIM~\cite{wang2004image}, and LPIPS~\cite{zhang2018unreasonable}. 
To further assess perceptual quality, we report no-reference metrics such as NIQE~\cite{mittal2012making}, CLIP-IQA~\cite{wang2023exploring}, MUSIQ~\cite{ke2021musiq}, and DOVER~\cite{wu2023exploring}. 
We also employ the flow warping error $E_{warp}^{*}$~(scaled by $10^{-3}$)~\cite{lai2018learning} to evaluate temporal consistency. 
For real-world~(VideoLQ) and AIGC~(AIGC60) datasets, where ground-truth HR videos are unavailable, we rely solely on no-reference metrics and $E_{warp}^{*}$ for evaluation.

\vspace{-0.4em}
\subsection{Comparison with Prior Works}
\vspace{-0.2em}
We compare our DUO-VSR with several recent state-of-the-art video super-resolution (VSR) models, including RealViformer~\cite{zhang2024realviformer}, VEnhancer~\cite{he2024venhancer}, MGLD~\cite{yang2024motion}, UAV~\cite{zhou2024upscale}, STAR~\cite{xie2025star}, DLoRAL~\cite{sun2025one}, DOVE~\cite{chen2025dove},  and SEEDVR2~\cite{wang2025seedvr2}.

\begin{figure}[!t]
\centering
\includegraphics[width=0.95\linewidth]{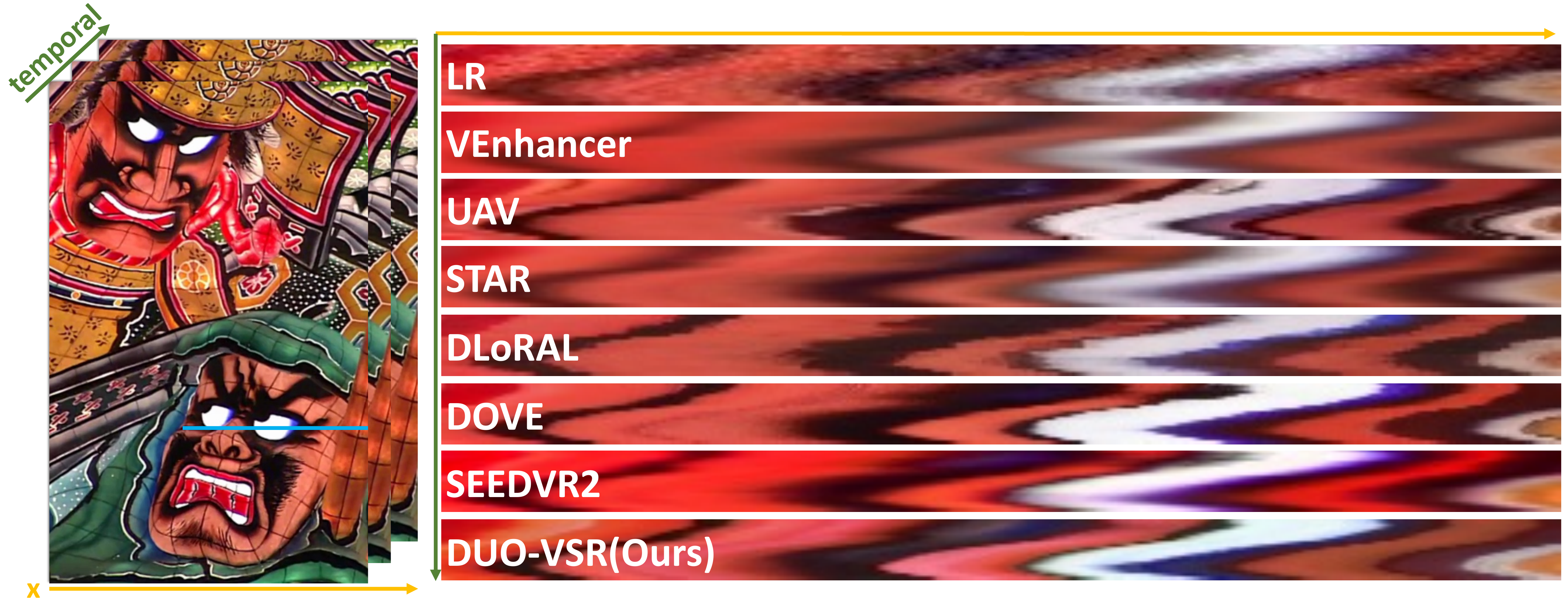}
\vspace{-1.em}
\caption{Comparison of temporal consistency. Extracted and stacked along the \textcolor[HTML]{4FACE8}{blue} line in the width–temporal plane.}
\vspace{-1.6em}
\label{fig:tem}
\end{figure}

\noindent\textbf{Qualitative Comparison.\ }Fig.~\ref{fig:teaser} and Fig.~\ref{fig:visualres} present qualitative comparisons with various methods on synthetic, real-world, and AIGC video datasets. DUO-VSR demonstrates strong capability in reconstructing realistic textures and structures under diverse and challenging degradations. For example, in Fig.~\ref{fig:visualres}, the first row shows that DUO-VSR successfully restores a visually convincing brick-wall pattern; in the second row, it reconstructs a clear human face even under severe degradation; and in the last row, it produces fine-grained, natural fur.
The temporal profiles visualized in Fig.~\ref{fig:tem} illustrate the comparison of temporal consistency. Under severely degraded LR inputs, existing methods tend to produce noticeable misalignment or blurring, whereas our DUO-VSR achieves a good balance between detail enhancement and temporal coherence.
More results are provided in the supplementary materials.

\noindent\textbf{Quantitative Comparison.\ }We present quantitative comparisons in Tab.~\ref{tab:quantitative-results} and Tab.~\ref{tab:params_comparison}. As can be seen, DUO-VSR consistently achieves the highest or near-highest scores on non-reference perceptual metrics such as NIQE and MUSIQ across all datasets, demonstrating its superior perceptual quality. In terms of fidelity metrics, our method attains performance comparable to competing approaches. Moreover, DUO-VSR exhibits highly stable and consistent results in temporal coherence~($E_{warp}^{*}$). In terms of efficiency, Tab.~\ref{tab:params_comparison} shows that DUO-VSR maintains low inference latency with a relatively small parameter scale. Compared with previous multi-step methods such as MGLD~\cite{yang2024motion}, it achieves near $90\times$ faster inference, and even compared with recent one-step approaches, its speed is generally more than $5\times$ higher. Overall, these comprehensive evaluations verify the effectiveness and superiority of our approach.

\begin{table}[h]
\centering
\vspace{-0.6em}
\caption{Inference efficiency comparison. Measured on a single GPU using a 21-frame $1920\times1080$ video. The model parameters are counted only for the generator part.}
\vspace{-0.8em}
\small     % 改成 small
\setlength{\tabcolsep}{4pt} % 稍微宽一点
\resizebox{\columnwidth}{!}{
\begin{tabular}{l|cccc|cccc}
\toprule
\textbf{Metric} & \textbf{UAV} & \textbf{MGLD} & \textbf{VEnh.} & \textbf{STAR} & \textbf{DOVE} & \textbf{DLoRAL} & \textbf{SeedVR2} & \textbf{DUO-VSR} \\
\midrule
\textbf{Step}        & 30 & 50 & 15 & 15 & 1 & 1 & 1 & 1 \\
\textbf{Time (s)}    & 382.1 & 956.7 & 404.5 & 200.4 & 66.7 & 76.6 & 89.7 & \textbf{11.3} \\
\textbf{Params (B)}  & \textbf{0.7} & 1.4 & 2.0 & 2.0 & 5.6 & 0.9 & 8.2 & 1.3 \\
\bottomrule
\end{tabular}}
\label{tab:params_comparison}
\vspace{-0.8em}
\end{table}

\vspace{-1.2em}
\subsection{Ablation Study}
\label{ablationstudy}
\vspace{-0.4em}
We conduct ablation studies to evaluate the contribution of each component and design choice, following the training configurations in Sec.~\ref{expsetting} and using the AIGC60 dataset. Further analysis is provided in the supplementary material.

\begin{figure}[!t]
\centering
\includegraphics[width=0.9\linewidth]{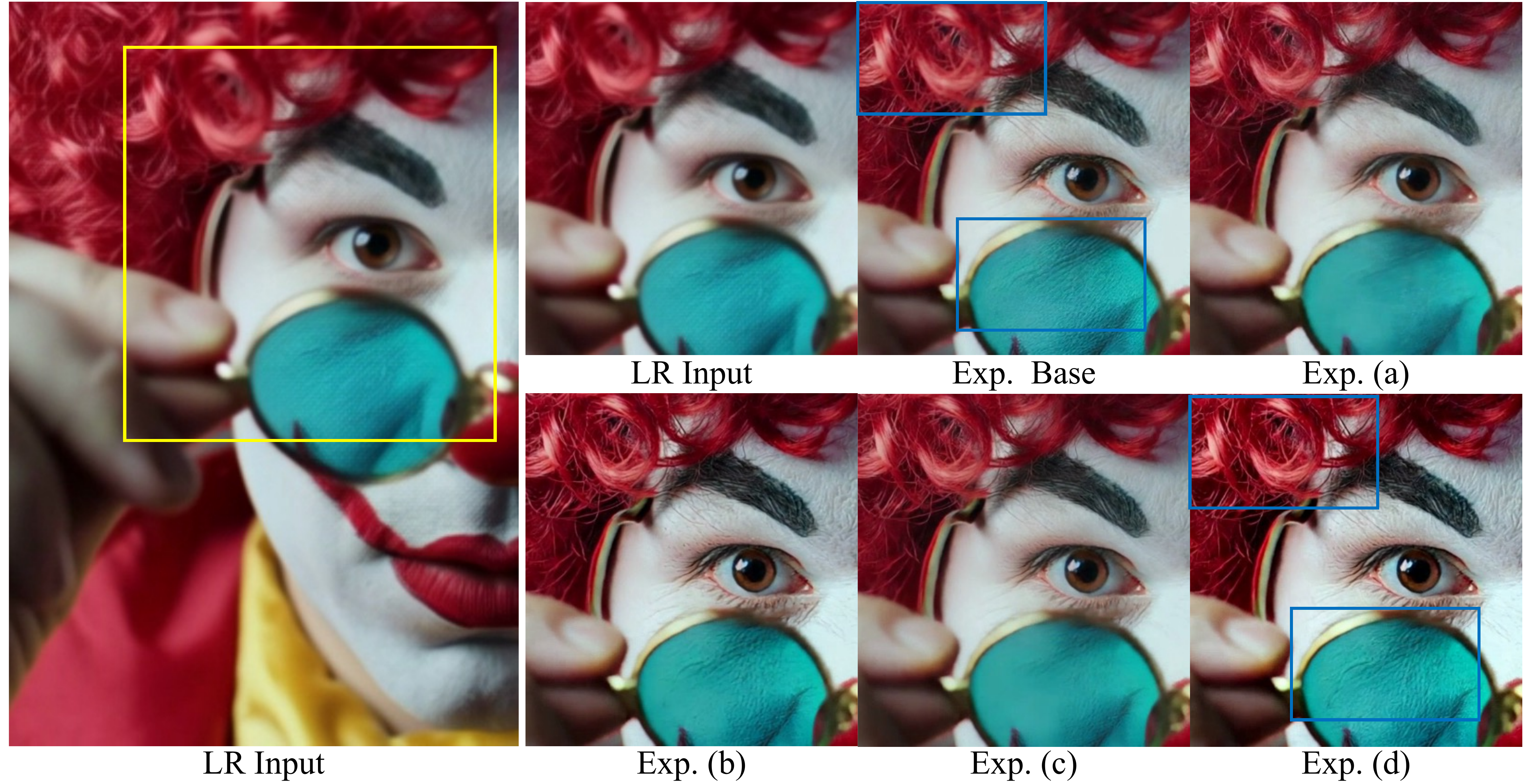}
\vspace{-0.8em}
\caption{Visual comparison of ablation on three stage distillation. Experiment indices refer to Tab.~\ref{abatable1}. Zoom in for details.}
\vspace{-1.8em}
\label{fig:abA}
\end{figure}

\noindent \textbf{Ablation on Three Stage Distillation.\ }
We present the ablation on the impact of our three-stage fine-tuning pipeline in Tab.~\ref{abatable1} and Fig.~\ref{fig:abA}. We report the performance of the base model with 50 inference steps (the first row highlighted in gray) and compare it with variants equipped with different fine-tuning stages (Exps.~(a)–(d), where $\checkmark$ indicates the inclusion of that stage). Comparing (a) and (b) shows that incorporating Dual-Stream Distillation (Stage II) notably improves the distilled model, benefiting from the RFS-GAN supervision derived from real-world videos and even surpassing the base model on perceptual metrics such as CLIPIQA and DOVER. Comparing (b) and (d) reveals that the Preference-Guided Refinement (Stage III) further enhances perceptual quality, demonstrating the effectiveness of preference-based alignment for human-perceived realism. Finally, the comparison between (c) and (d) highlights that the Trajectory-Preserving Distillation (Stage I) provides a strong initialization that stabilizes subsequent training and contributes to consistent quality improvements.

\begin{table}[h]
    \centering
    \vspace{-1.0em}
    \caption{Ablation on Three Stage Distillation.}
    \vspace{-0.9em}
    \begin{tabularx}{\linewidth}{c| X X X|X X X X}
        \toprule
        \multirow{2}{*}{Exp.} & \multicolumn{3}{c|}{{Variants}} & \multicolumn{4}{c}{{Metrics}} \\ 
        & \scriptsize{I} & \scriptsize{II} & \scriptsize{III} & \scriptsize{NIQE} & \scriptsize{MUSIQ} & \scriptsize{CLIPIQA} & \scriptsize{DOVER} \\ \hline

        \rowcolor{gray!10}
        Base & & & & \textbf{4.31} & 63.46 & 0.4712 &87.98 \\
        (a) &\checkmark &  & & 5.45 & 58.97 & 0.408 & 86.49 \\
        (b) &\checkmark &\checkmark & & 4.64& 63.36 & 0.487 & 88.01 \\
        (c) & &\checkmark &\checkmark &5.11 & 60.22 & 0.423 & 87.63 \\
        (d) &\checkmark &\checkmark &\checkmark & 4.42 & \textbf{63.68} & \textbf{0.489} & \textbf{88.15} \\
        \bottomrule
    \end{tabularx}
    \label{abatable1}
    \vspace{-1.em}
\end{table}

\begin{figure}[!t]
\centering
\includegraphics[width=0.9\linewidth]{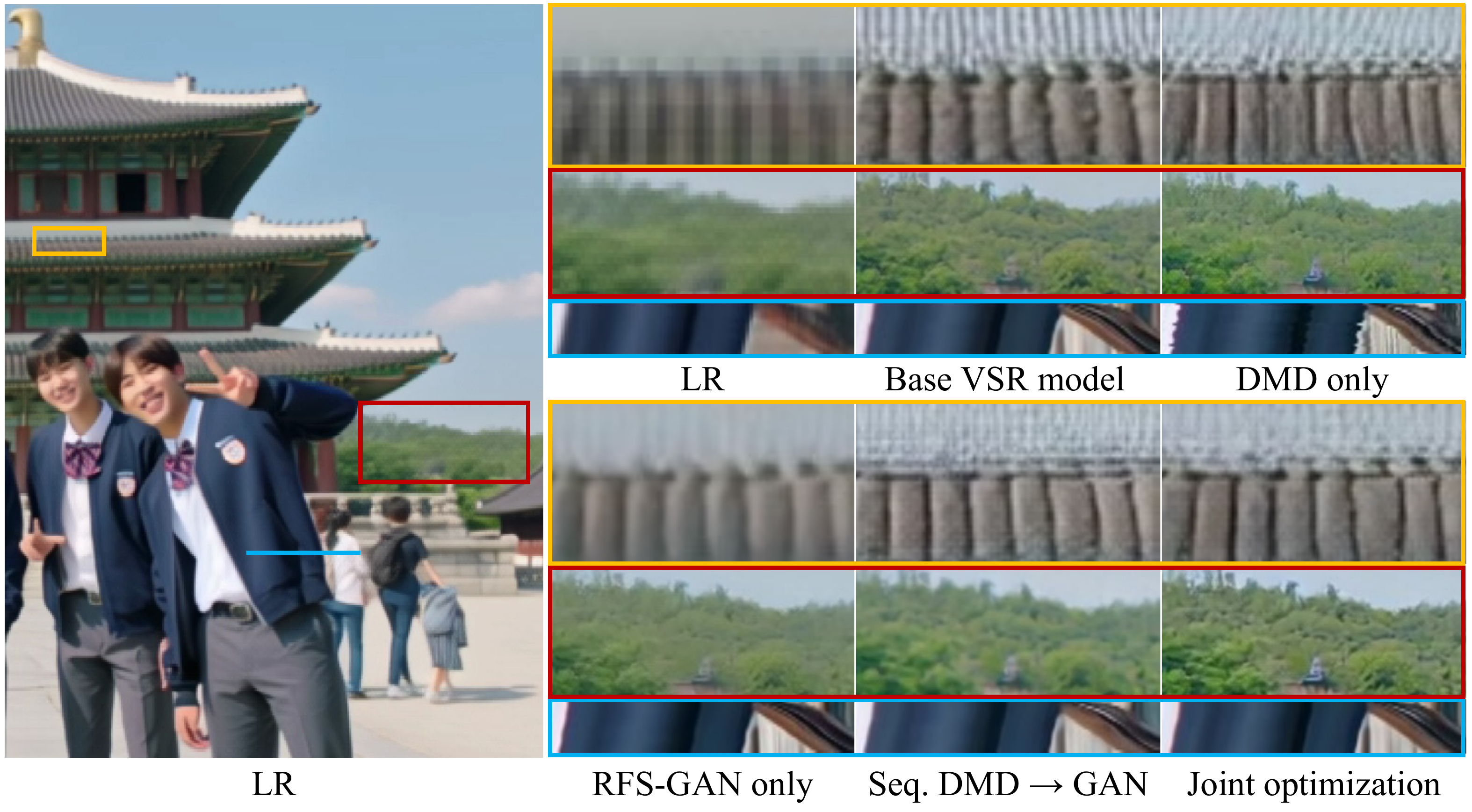}
\vspace{-0.8em}
\caption{Visual comparison of ablation on Dual-Stream Distillation Strategy. The \textcolor[HTML]{F4C242}{orange} and \textcolor[HTML]{AF2519}{red} boxes show spatial comparison in the LR. The \textcolor[HTML]{4FACE8}{blue} box shows the temporal profile along the blue line in the LR. Zoom in for details.}
\vspace{-2.0em}
\label{fig:abB}
\end{figure}

\noindent \textbf{Ablation on Dual-Stream Distillation Strategy.\ }
We further analyze the Dual-Stream Distillation strategy in Tab.~\ref{tab:dualstream} and Fig.~\ref{fig:abB}, where we separately examine its effectiveness from the component and optimization perspectives.
At the component level, we compare the effects of using DMD or RFS-GAN alone against their combination. While each individual branch provides moderate improvements compared to using only Stage I, the Dual-Stream configuration enables them to play complementary roles, achieving notably better performance across all perceptual metrics.
As shown in Fig.~\ref{fig:abB}, while RFS-GAN alone does not enhance textures as effectively as DMD (e.g., plants in the red box), its inclusion mitigates quality degradation or insufficient supervision from DMD alone (e.g., tiles in the orange box, temporal profile in the blue box), improving artifacts and temporal consistency.
At the optimization level, we compare our joint optimization scheme with the sequential strategy adopted in DMD2, which first distills with DMD and then fine-tunes with GAN supervision. The results show that joint optimization allows the two objectives to interact more effectively during training, leading to stronger mutual reinforcement and the best overall performance.

\begin{table}[h]
\centering
\vspace{-1.1em}
\caption{Ablation on Dual-Stream Distillation Strategy. ``Joint'' and ``Seq.'' denote different optimization schemes.}
\vspace{-0.9em}
\setlength{\tabcolsep}{3pt}
\small
\begin{tabularx}{\linewidth}{>{\raggedright\arraybackslash}X|cccc}
\toprule
Setting & \scriptsize{NIQE} & \scriptsize{MUSIQ} & \scriptsize{CLIPIQA} & \scriptsize{DOVER} \\
\midrule
%Base VSR model (50 steps) & 4.31 & 63.46& 0.471& 87.98 \\ \midrule
\rowcolor{gray!10}\multicolumn{5}{@{}l@{}}{\textit{Component-level}} \\
DMD only & 4.99 & 61.46 & 0.432 & 87.38 \\
RFS-GAN only & 5.32 & 62.64 & 0.427 & 87.53 \\
Dual-Stream (Joint) & \textbf{4.42} & \textbf{63.68} & \textbf{0.489} & \textbf{88.15} \\
\midrule
\rowcolor{gray!10}\multicolumn{5}{@{}l@{}}{\textit{Optimization-level}} \\
Sequential DMD$\rightarrow$GAN (Seq.) & 5.17 & 62.76 & 0.419 & 87.67 \\
Joint optimization (ours) & \textbf{4.42} & \textbf{63.68} & \textbf{0.489} & \textbf{88.15}  \\
\bottomrule
\end{tabularx}
\vspace{-1.8em}
\label{tab:dualstream}
\end{table}

%% file: sec/5_conclusion.tex
\vspace{-0.5em}
\section{Conclusion}
\vspace{-0.6em}
In this paper, we identified that directly applying distribution matching distillation (DMD) to one-step video super-resolution suffers from training instability, degraded supervision from the real score model, and insufficient guidance toward real HR videos.
To address these issues, we proposed DUO-VSR, a three-stage framework built upon a Dual-Stream Distillation Strategy that integrates DMD with Real–Fake Score Feature GAN for stable and comprehensive supervision.
Through Progressive Guided Distillation Initialization, Dual-Stream Distillation, and Preference-Guided Refinement, DUO-VSR effectively stabilizes optimization, enhances supervision, and aligns perceptual quality preferences.
Our findings reveal that combining distribution matching and adversarial supervision provides an effective path toward efficient, high-fidelity one-step VSR.

%% file: sec/6_ack.tex
\vspace{-1em}
\section{Acknowledgements}
\vspace{-0.6em}
This work was partly supported by the National Natural Science Foundation of China (Grant No. 62502169)

%% file: sec/X_suppl.tex
\clearpage
\setcounter{page}{1}
\maketitlesupplementary

\section{Further Implementation Details}
% \vspace{-0.8em}
\subsection{Algorithm for Dual-Stream Distillation}
% \vspace{-0.4em}
The detailed procedure of the dual-stream distillation strategy is outlined in Algorithm~\ref{alg:dual_stream_simple_2e_fixed}, comprising interleaved Auxiliary and Student updates. In our implementation, we set the update interval $N=3$ by default.

\subsection{Construction of Preference Dataset}
\begin{figure}[t]
\centering
\includegraphics[width=0.99\linewidth]{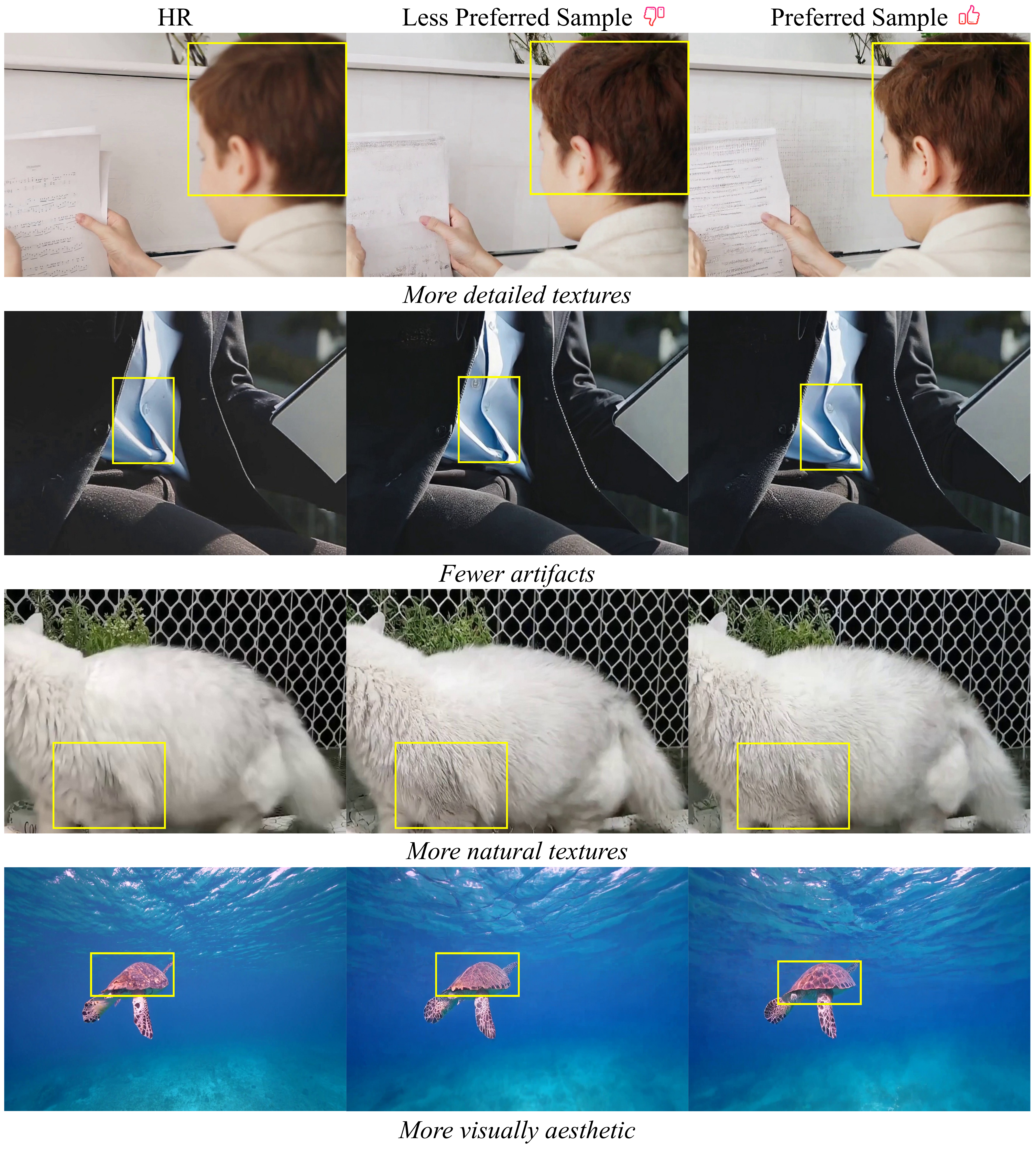}
\vspace{-0.8em}
\caption{Examples of preferred and less-preferred samples in the constructed preference dataset. Zoom in for details.}
\vspace{-1.6em}
\label{fig:prefer}
\end{figure}

In the preference-guided refinement stage, we construct a preference dataset for Direct Preference Optimization.
Specifically, for each LR video, we generate five candidate reconstructions using the second-stage model.
We then evaluate these candidates using the LPIPS~\cite{zhang2018unreasonable}, MUSIQ~\cite{ke2021musiq} and DOVER~\cite{wu2023exploring} metrics and rank them according to their combined quality scores.
The highest-scoring output is selected as the preferred sample, while the lowest-scoring one serves as the less preferred sample.
As illustrated in Fig.~\ref{fig:prefer}, the preferred samples typically exhibit richer, more natural, and aesthetically pleasing textures.
In total, we construct 2000 preference pairs for fine-tuning.

\begin{algorithm}[t]
\caption{Dual-Stream Distillation Strategy}
\label{alg:dual_stream_simple_2e_fixed}
\small
\KwIn{
Frozen Real Score model \(\theta_{\mathrm{R}}\); trainable Fake Score model \(\theta_{\mathrm{F}}\); student \(\theta_{\mathrm{S}}\); discriminator heads \(H_{\phi}\); loss weights \(\lambda_{\mathrm{DMD}}, \lambda_{\mathrm{GAN}}, \lambda_{\mathrm{FM}}\); interval \(N\).
}
\While{not converged}{
  \For{\(i \leftarrow 1\) \KwTo \(N\)}{
    \tcc{Auxiliary update}
    Sample \((\boldsymbol{z}^{LR}, \boldsymbol{z}^{HR}, \boldsymbol{c})\), \(t\), \(\epsilon\)\;
    \(\hat{\boldsymbol{z}}^{S}_0 \leftarrow \epsilon - \boldsymbol{v}_{\theta_{\mathrm{S}}}(\epsilon, t, \boldsymbol{z}^{LR}, \boldsymbol{c})\)\;
    \(\hat{\boldsymbol{z}}^{S}_t \leftarrow q_t\!\big(\hat{\boldsymbol{z}}^{S}_0\big)\), \(\;\boldsymbol{z}^{HR}_t \leftarrow q_t\!\big(\boldsymbol{z}^{HR}\big)\)\;

    \tcp{Diffusion loss for \(\theta_{\mathrm{F}}\)}
    Compute target \(\boldsymbol{v}\) at \(\big(\hat{\boldsymbol{z}}^{S}_t, t, \boldsymbol{z}^{LR}, \boldsymbol{c}\big)\)\;
    \(\mathcal{L}_{\mathrm{Diff}} \leftarrow \big\|\boldsymbol{v}_{\theta_{\mathrm{F}}}(\hat{\boldsymbol{z}}^{S}_t, t, \boldsymbol{z}^{LR}, \boldsymbol{c}) - \boldsymbol{v}\big\|^2\)\;

    \tcp{GAN discriminator loss for \(\phi\) with stop\_grad backbones}
    
    \(\mathbf{h}^{S} \leftarrow 
    \mathrm{concat}\!\big(\mathrm{Feat}_{\theta_{\mathrm{R}}}(\hat{\boldsymbol{z}}^{S}_t),
    \mathrm{Feat}_{\theta_{\mathrm{F}}}(\hat{\boldsymbol{z}}^{S}_t)\big)\)\;
    
    \(\mathbf{h}^{HR} \leftarrow \mathrm{concat}\!\big(\mathrm{Feat}_{\theta_{\mathrm{R}}}(\boldsymbol{z}^{HR}_t),\, \mathrm{Feat}_{\theta_{\mathrm{F}}}(\boldsymbol{z}^{HR}_t)\big)\)\;

    \(D_S \leftarrow H_{\phi}\!\big(\operatorname{sg}[\mathbf{h}^{S}]\big)\)\;
    \(D_{HR} \leftarrow H_{\phi}\!\big(\operatorname{sg}[\mathbf{h}^{HR}]\big)\)\;
    {\scriptsize\(\mathcal{L}_{\mathrm{D}} \leftarrow 
    \mathbb{E}[\max(0,1-D_{HR})] + \mathbb{E}[\max(0,1+D_S)]\)}\;

    Update \(\theta_{\mathrm{F}}\) by descending \(\nabla_{\theta_{\mathrm{F}}}\mathcal{L}_{\mathrm{Diff}}\)\;
    Update \(\phi\) by descending \(\nabla_{\phi}\mathcal{L}_{\mathrm{D}}\)\;
  }

  \tcc{Student update (after every \(N\) Auxiliary steps)}
  Sample \((\boldsymbol{z}^{LR}, \boldsymbol{z}^{HR}, \boldsymbol{c})\), \(t\), \(\epsilon\)\;
  \(\hat{\boldsymbol{z}}^{S}_0 \leftarrow \epsilon - \boldsymbol{v}_{\theta_{\mathrm{S}}}(\epsilon, t, \boldsymbol{z}^{LR}, \boldsymbol{c})\)\;
  \(\hat{\boldsymbol{z}}^{S}_t \leftarrow q_t\!\big(\hat{\boldsymbol{z}}^{S}_0\big)\), \(\;\boldsymbol{z}^{HR}_t \leftarrow q_t\!\big(\boldsymbol{z}^{HR}\big)\)\;

  \tcp{DMD loss}
  \(\hat{\boldsymbol{z}}^{R}_0 \leftarrow \hat{\boldsymbol{z}}^{R}_0(\hat{\boldsymbol{z}}^{S}_t; \theta_{\mathrm{R}})\), \quad
  \(\hat{\boldsymbol{z}}^{F}_0 \leftarrow \hat{\boldsymbol{z}}^{F}_0(\hat{\boldsymbol{z}}^{S}_t; \theta_{\mathrm{F}})\)\;
  \(\text{Grad} \leftarrow \dfrac{\hat{\boldsymbol{z}}^{F}_0 - \hat{\boldsymbol{z}}^{R}_0}{\operatorname{mean}\!\big(\operatorname{abs}(\hat{\boldsymbol{z}}^{S}_0 - \hat{\boldsymbol{z}}^{R}_0)\big)}\)\;
  \(\mathcal{L}_{\mathrm{DMD}} \leftarrow \big\|\hat{\boldsymbol{z}}^{S}_0 - \operatorname{sg}[\hat{\boldsymbol{z}}^{S}_0 - \text{Grad}]\big\|^2\)\;

  \tcp{GAN generator loss}
  \(\mathbf{h}^{S} \leftarrow \mathrm{concat}\!\big(\mathrm{Feat}_{\theta_{\mathrm{R}}}(\hat{\boldsymbol{z}}^{S}_t),\, \mathrm{Feat}_{\theta_{\mathrm{F}}}(\hat{\boldsymbol{z}}^{S}_t)\big)\)\;
  \(D(\hat{\boldsymbol{z}}^{S}_t) \leftarrow H_{\phi}\!\big(\operatorname{sg}[\mathbf{h}^{S}]\big)\)\;
  \(\mathcal{L}_{\mathrm{G}} \leftarrow -\mathbb{E}\!\big[D(\hat{\boldsymbol{z}}^{S}_t)\big]\)\;

  \tcp{Feature matching loss}
  \(\mathbf{h}^{HR} \leftarrow \mathrm{concat}\!\big(\mathrm{Feat}_{\theta_{\mathrm{R}}}(\boldsymbol{z}^{HR}_t),\, \mathrm{Feat}_{\theta_{\mathrm{F}}}(\boldsymbol{z}^{HR}_t)\big)\)\;
 
  \(\mathcal{L}_{\mathrm{FM}} \leftarrow \big\|\mathbf{h}^{S} - \mathbf{h}^{HR} \big\|^2\)\;
  
  \(\mathcal{L}_{\mathrm{S}} \leftarrow \lambda_{\mathrm{DMD}}\mathcal{L}_{\mathrm{DMD}} + \lambda_{\mathrm{GAN}}\mathcal{L}_{\mathrm{G}} + \lambda_{\mathrm{FM}}\mathcal{L}_{\mathrm{FM}}\)\;
  Update \(\theta_{\mathrm{S}}\) by descending \(\nabla_{\theta_{\mathrm{S}}}\mathcal{L}_{\mathrm{S}}\)\;
}
\end{algorithm}

\vspace{-0.2em}
\section{Further Discussions and Ablation Analyses.}
\vspace{-0.2em}
In the ablation study presented in Sec.~\ref{ablationstudy} of the main text, we analyzed the effectiveness of the three-stage distillation framework and the two branches in the Dual-Stream Distillation, namely the DMD stream and the RFS-GAN stream, along with the exploration of different optimization strategies. In this section, we provide additional discussions on the design and training of the RFS-GAN.

\begin{figure}[t]
\centering
\includegraphics[width=0.99\linewidth]{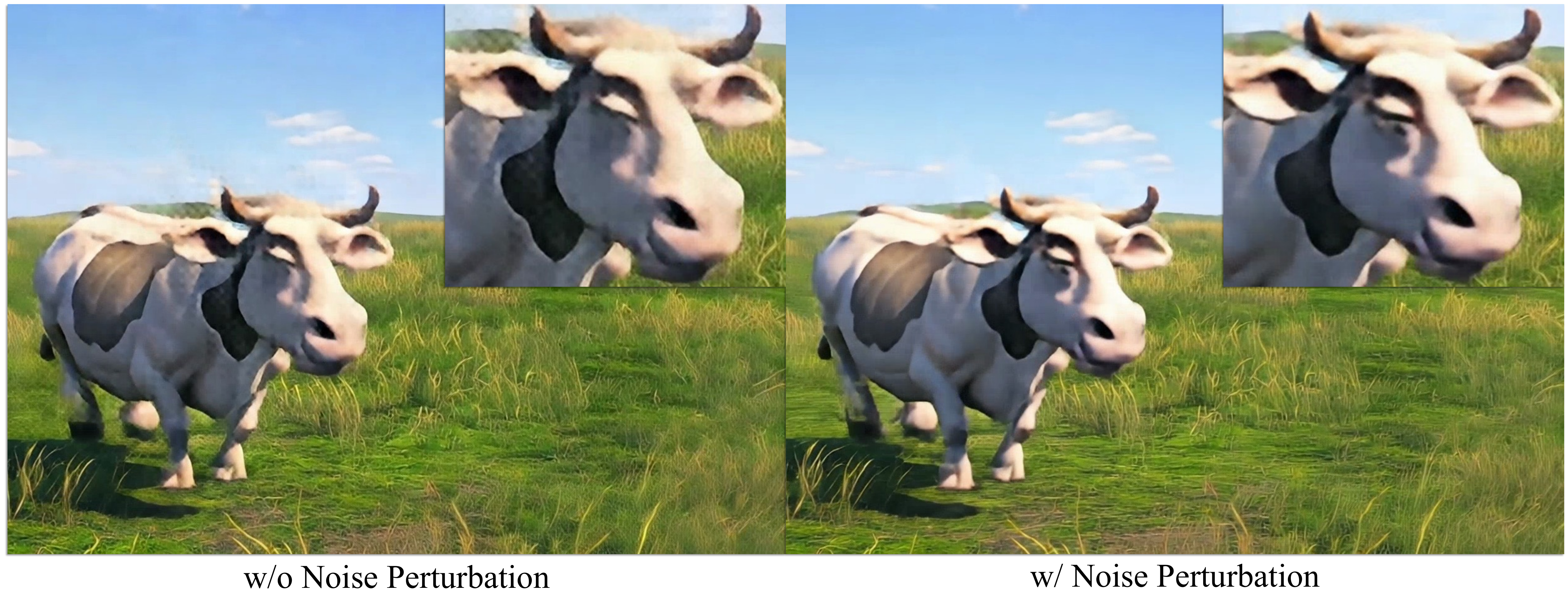}
\vspace{-0.4em}
\caption{Noise-perturbed samples stabilize adversarial training and suppress artifacts. Zoom in for details.}
\vspace{-1.8em}
\label{fig:abC}
\end{figure}

\noindent\textbf{Noise-Perturbed Sample Input in RFS-GAN.} Different from SeedVR2~\cite{wang2025seedvr2}, which directly feeds the clean outputs of the student into the discriminator, we observe that such a design often leads to training instability and occasionally produces grid-like artifacts, as shown in Fig.~\ref{fig:abC}. We hypothesize that this instability stems from a discriminator–generator imbalance, where an overly strong discriminator can easily distinguish real samples from fake ones. Inspired by the perturbation strategy in DMD~\cite{yin2024one}, which intentionally blurs the boundary between real and fake data distributions, we similarly add random noise with varying intensity to both real and fake inputs of the discriminator. This modification effectively stabilizes the adversarial learning while preserving its enhancement effect.

Furthermore, using noisy real and fake samples enables sharing the intermediate features from real and fake score computation for the GAN loss calculation, requiring only an additional extraction of features from real samples and thus reducing the number of forward passes.

\begin{figure}[t]
\centering
\includegraphics[width=0.99\linewidth]{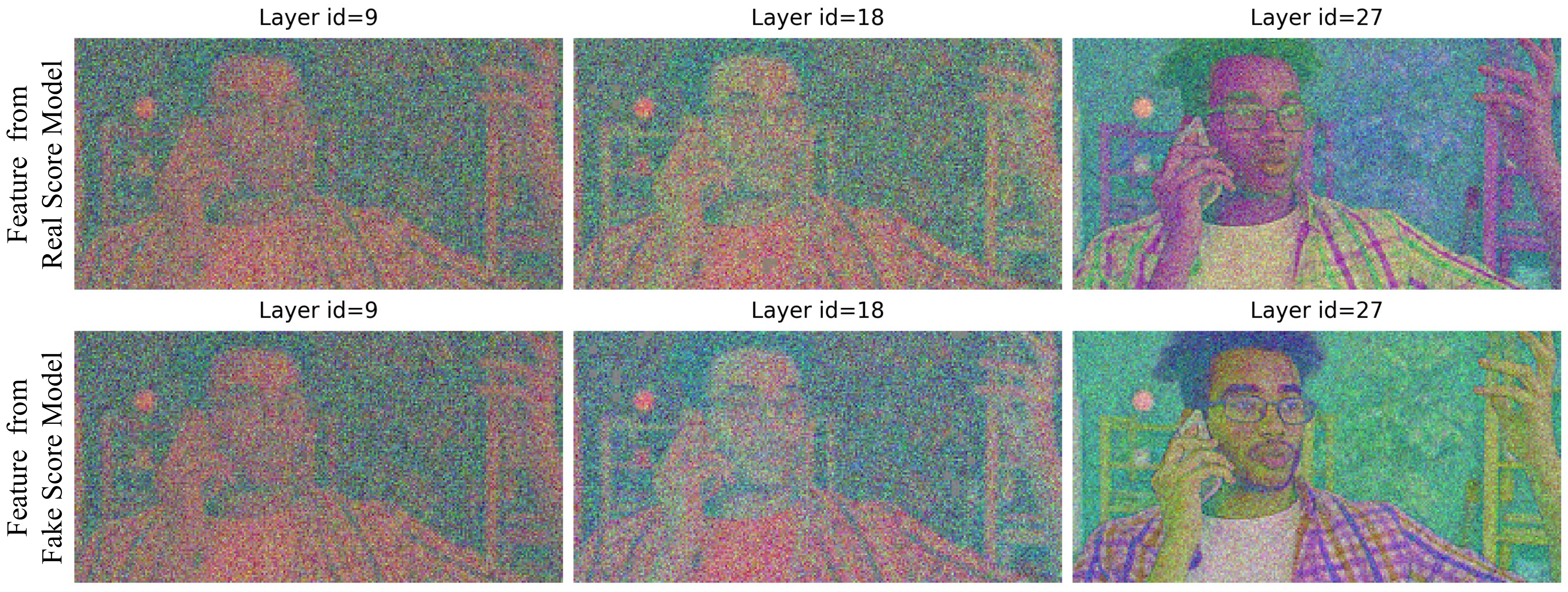}
\vspace{-0.8em}
\caption{Discriminator features from the real and fake score models used for the RFS-GAN loss computation, reduced to three dimensions via t-SNE~\cite{maaten2008visualizing} for visualization.}
\vspace{-1.2em}
\label{fig:abD}
\end{figure}

\noindent\textbf{Cross-Model and Multi-Layer Feature in RFS-GAN} In RFS-GAN, both the real score model and the fake score model are employed as the backbones of the discriminator. As illustrated in Fig.~\ref{fig:abD}, intermediate representations are extracted from the 9th, 18th, and 27th layers of the DiT architecture (consisting of 30 layers in total). RFS-GAN effectively integrates shallow features that capture structural and semantic information with deeper representations that encode richer and more fine-grained details. Furthermore, the two score models are optimized over distinct data distributions: the real score model is intrinsically aligned with the real (teacher) distribution, providing high-quality discriminative guidance, whereas the fake score model dynamically reflects the evolving distribution of the student. The complementarity between these two models substantially enhances the representational capacity of the discriminator, thereby delivering stronger and more reliable gradient feedback to the student model.

Ablation studies are performed on the second-stage model to assess the effectiveness of discriminator features extracted from the real and fake score models. As shown in Tab.~\ref{tab:rfsgan}, the discriminator that combines the real and fake score models achieves the best performance in perceptual metrics, demonstrating the effectiveness of RFS-GAN.

\begin{table}[t]
\centering
\caption{Ablation study on the discriminator design of RFS-GAN.}
\vspace{-0.6em}
\scriptsize
\setlength{\tabcolsep}{6pt}
\resizebox{\columnwidth}{!}{
\begin{tabular}{lcccc}
\toprule
Method & NIQE$\downarrow$ & MUSIQ $\uparrow$& CLIPIQA $\uparrow$& DOVER $\uparrow$\\
\midrule
Real Score Model only  & 4.71 & 62.79 & 0.456 &87.95  \\
Fake Score Model only  & 4.98 & 62.98 & 0.475 & 87.76 \\
RFS-GAN  & \textbf{4.64} & \textbf{63.36} &  \textbf{0.487} & \textbf{88.01} \\
\bottomrule
\end{tabular}
}
\vspace{-0.8em}
\label{tab:rfsgan}
\end{table}

\section{Additional Evaluation Results}

\subsection{Additional Visual Comparisons}
\begin{figure*}[!t]
\centering
\includegraphics[width=0.88\linewidth]{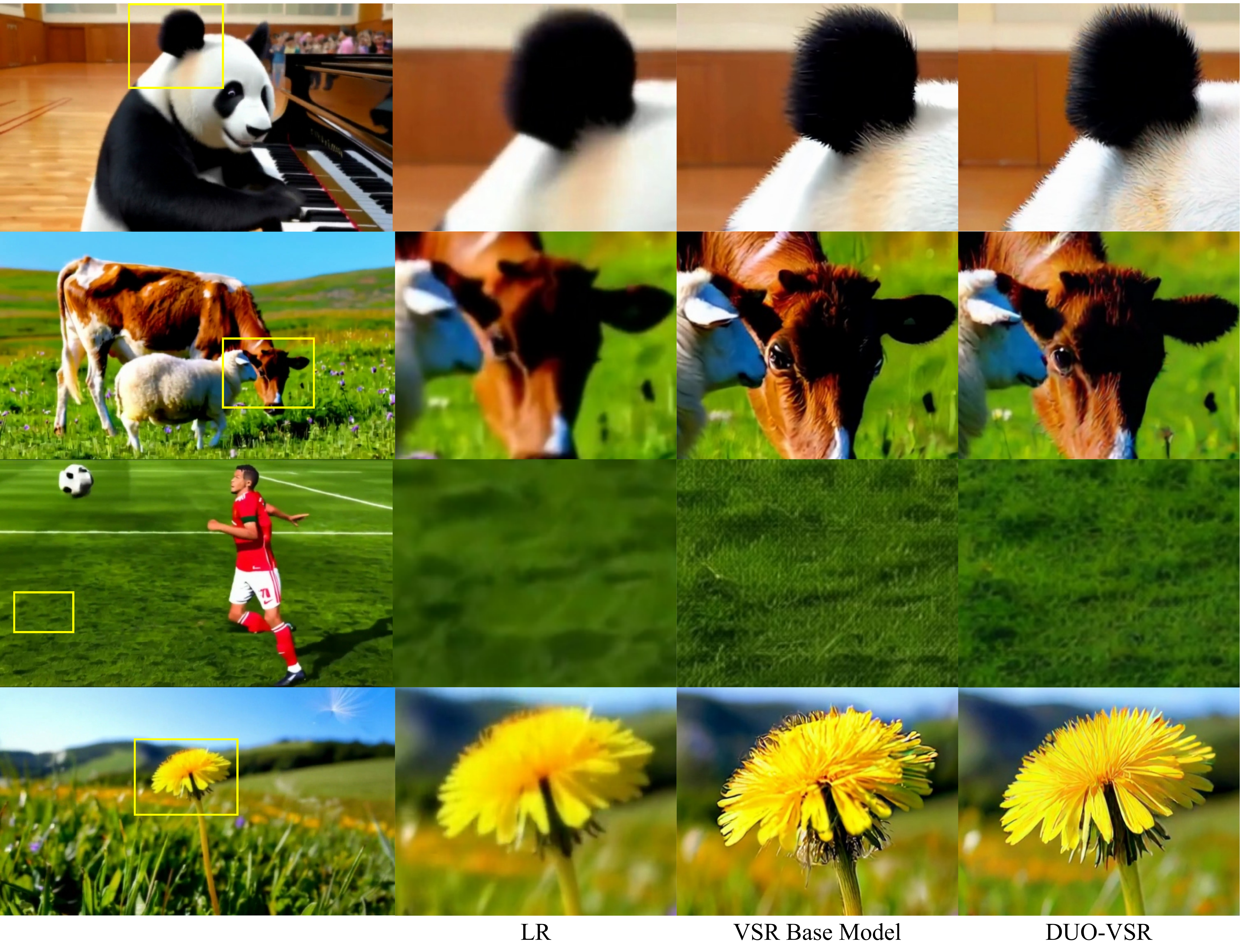}
\vspace{-0.6em}
\caption{Visual comparison with base VSR model. Zoom in for details.}
\vspace{-0.4em}
\label{fig:basecmp}
\end{figure*}

\noindent\textbf{Comparison with the base model.} We first compare DUO-VSR with its base model to examine the effectiveness of the distillation framework, as shown in the Fig.~\ref{fig:basecmp}. The results indicate that our method achieves a comparable ability to generate textures (first row), while producing more natural and visually coherent details (third and fourth rows).

\noindent\textbf{Comparison with other methods.} We present additional visual quality comparisons with VEnhancer~\cite{he2024venhancer}, UAV~\cite{zhou2024upscale}, STAR~\cite{xie2025star}, DLoRAL~\cite{sun2025one}, DOVE~\cite{chen2025dove},  and SEEDVR2~\cite{wang2025seedvr2} in Fig.~\ref{fig:supp}. These results further demonstrate the advantages of our method when dealing with challenging regions that involve fine textures.

% \textbf{More video results can be found in the demo video provided in the supplementary material.}

\subsection{Discussion of Concurrent Works}
%We note that several concurrent works have also explored DMD for efficient video super-resolution. 
We note that several concurrent works~\cite{zhang2025infvsr, zhuang2025flashvsr,guo2025towards} have explored efficient video super-resolution, some of which also employ DMD for one-step inference.
Both InfVSR~\cite{zhang2025infvsr} and FlashVSR~\cite{zhuang2025flashvsr} adopt DMD and causal DiT architectures to achieve one-step streaming VSR, focusing primarily on reformulating full-sequence diffusion into a causal structure, where DMD mainly serves as a step-distillation mechanism. 
Earlier, UltraVSR~\cite{liu2025ultravsr} also employs distribution matching distillation to facilitate one-step VSR, but focuses on degradation-aware scheduling and leverages an image diffusion backbone (extended Stable Diffusion~\cite{rombach2022high} for VSR). 
In contrast, our DUO-VSR takes an orthogonal perspective by revisiting the intrinsic limitations of DMD in VSR and introducing an effective dual-stream distillation strategy to mitigate them. This design offers a complementary pathway that could potentially be integrated with existing DMD-based frameworks to further enhance their robustness and visual quality.

\begin{figure*}[!t]
\centering
\includegraphics[width=0.99\linewidth]{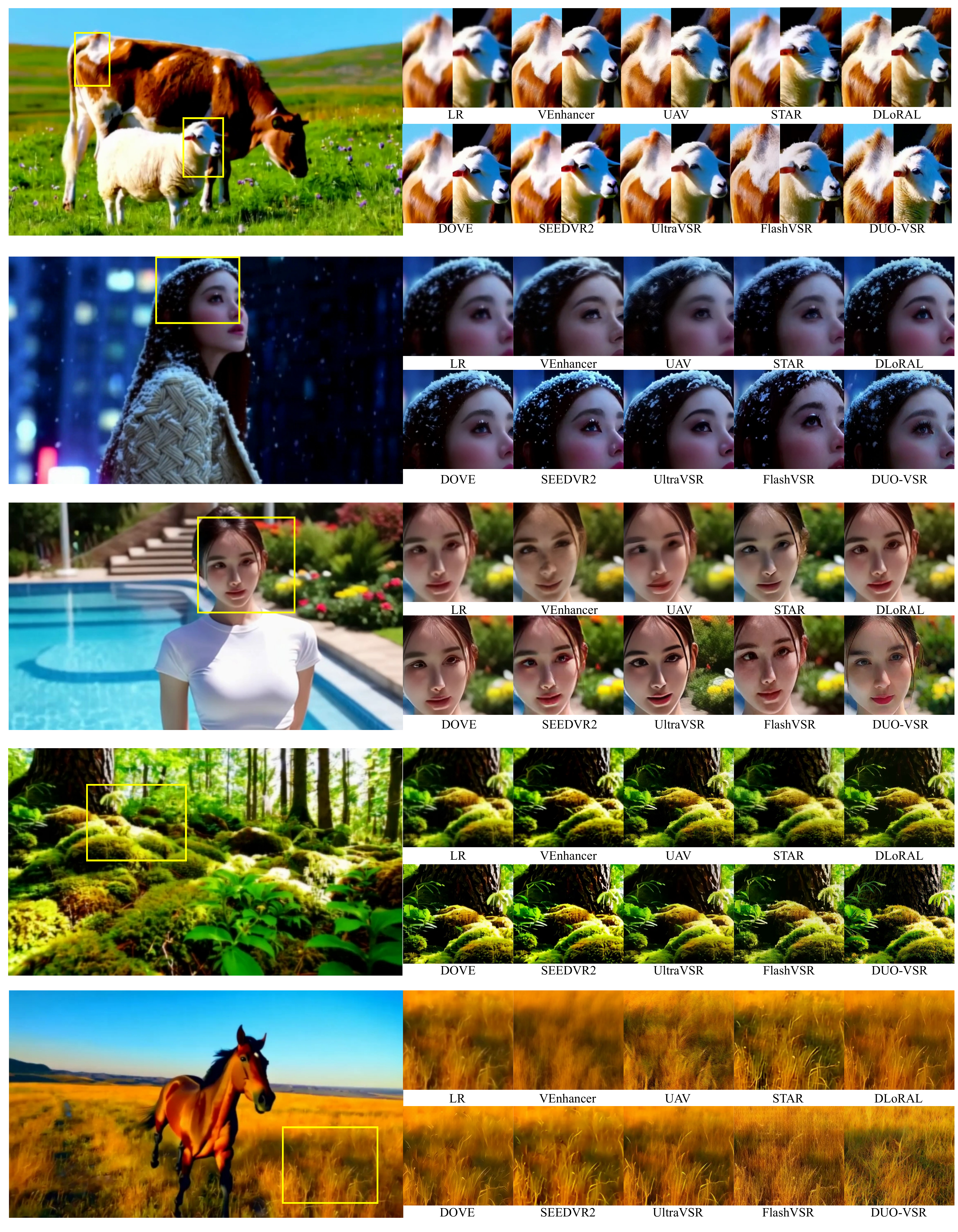}
\vspace{-1.0em}
\caption{Visual comparison of different VSR methods. DUO-VSR consistently reconstructs finer textures. Zoom in for details.}
\vspace{-1.4em}
\label{fig:supp}
\end{figure*}

Recently, both FlashVSR~\cite{zhuang2025flashvsr} and UltraVSR~\cite{liu2025ultravsr} have made their official implementations publicly available, and we include comparative results in this supplementary material. To ensure a fair comparison in terms of performance and quality, we use FlashVSR-Full for evaluation. As shown in Fig.~\ref{fig:supp}, our method produces more realistic and natural details than FlashVSR and UltraVSR. Specifically, in the first case, DUO-VSR reconstructs finer and smoother fur textures on the fox; in the second case, the woman’s eyebrows and eyes appear more natural; and in the fourth case, the wheat spikes exhibit more faithful and visually convincing structures. Tab.~\ref{tab:aigc60_results} presents a quantitative comparison between DUO-VSR and these two methods on the AIGC60 dataset. 
It can be seen that DUO-VSR achieves superior performance in perceptual metrics while exhibiting comparable inference efficiency to FlashVSR-Full.

\begin{table}[t]
\centering
\caption{Quantitative comparison on the AIGC60 dataset.}
\vspace{-0.8em}
\small
\setlength{\tabcolsep}{6pt}
\resizebox{\columnwidth}{!}{
\begin{tabular}{lccc}
\toprule
Metric & UltraVSR & FlashVSR & DUO-VSR \\
\midrule
NIQE $\downarrow$ & 5.58 & \underline{4.67} & \textbf{4.42} \\
MUSIQ $\uparrow$ & 58.23 & \underline{63.11} & \textbf{63.68} \\
CLIP-IQA $\uparrow$ & 0.4434 & \underline{0.4690} & \textbf{0.4886} \\
DOVER $\uparrow$ & 86.45 & \underline{87.49} & \textbf{88.15} \\
$E_{warp}^{*}\downarrow$ & \underline{1.54} & 1.76 & \textbf{1.08} \\
\hline
Time (s) & 126.5 & \textbf{10.7} & \underline{11.3} \\
Params (B) & \underline{1.9} & \textbf{1.3} & \textbf{1.3} \\
\bottomrule
\end{tabular}}
\label{tab:aigc60_results}
\vspace{-2em}
\end{table}

\subsection{User Study}
Following APT~\cite{lin2025diffusion} and SeedVR2~\cite{wang2025seedvr2}, we conducted a blind user study using the GSB test to more comprehensively assess the subjective visual quality of our method. Specifically, the preference score is computed as $\frac{G-B}{(G + B + S)}$, where G denotes the number of samples judged as good, B as bad, and S as similar. The score ranges from -100\% to 100\%, with 0\% indicating equal performance. We randomly selected 30 samples from the VideoLQ and AIGC60 datasets. The evaluation primarily compared our approach with recent one-step video super-resolution methods, including SeedVR2-7B~\cite{wang2025seedvr2}, DOVE~\cite{chen2025dove}, DLoRAL~\cite{sun2025one}, UltraVSR~\cite{liu2025ultravsr}, and FlashVSR-Full~\cite{zhuang2025flashvsr}. Participants rated three aspects: visual fidelity, visual quality, and overall quality. Twenty researchers with computer vision backgrounds took part in the evaluation. As shown in Tab.~\ref{tab:userstudy}, DUO-VSR achieves higher subjective preference scores than previous methods.

\begin{table}[t]
\centering
\caption{Blind user study results based on GSB test.}
\vspace{-0.8em}
\small
\setlength{\tabcolsep}{6pt}
\resizebox{\columnwidth}{!}{
\begin{tabular}{lccc}
\toprule
Method & Overall Quality & Visual Fidelity & Visual Quality \\
\midrule
DUO-VSR & 0\% & 0\% & 0\% \\
Our Base VSR model & -1.3\% & -3.7\% & +2.3\% \\
\hline
SeedVR2-7B & -32.7\% & -13.3\% & -39.2\% \\
DOVE & -29.3\% & -8.0\% & -36.2\% \\
DLoRAL & -34.0\% & -10.8\% & -37.8\% \\
UltraVSR & -39.8\% & -16.7\% & -43.3\% \\
FlashVSR-Full & -25.5\% & -6.7\% & -28.2\% \\
\bottomrule
\end{tabular}}
\label{tab:userstudy}
\vspace{-1em}
\end{table}

\section{Limitations and Future Work}
\noindent\textbf{Limitations.\ }Despite the strong efficiency and perceptual quality achieved by our one-step framework, several limitations remain. Since our method is trained in the latent space, the underlying VAE applies an aggressive spatiotemporal compression ($8\times$ spatial and $4\times$ temporal), which can hinder the reconstruction of extremely fine-grained details such as tiny text. In addition, the video VAE becomes the dominant computational bottleneck during inference, accounting for more than 90\% of the total runtime.

\noindent\textbf{Future Work.\ }In future work, we plan to explore more efficient or task-specific video VAEs that not only preserve high-frequency details and temporal coherence but also significantly accelerate the decoding process, thereby reducing the overall inference latency of our one-step framework.